\documentclass{article} 
\ifdefined\pdfsuppressptexinfo\pdfsuppressptexinfo=-1\fi
\usepackage{iclr2027_conference,times}
\definecolor{rowgray}{HTML}{DFDFDF}

\usepackage{amsmath,amsfonts,bm}

\def\eqref#1{equation~\ref{#1}}

\def\1{\bm{1}}

\def\ra{{\textnormal{a}}}

\def\rx{{\textnormal{x}}}

\def\rva{{\mathbf{a}}}

\def\erva{{\textnormal{a}}}

\def\ervx{{\textnormal{x}}}

\def\rmA{{\mathbf{A}}}

\def\vmu{{\bm{\mu}}}
\def\vtheta{{\bm{\theta}}}
\def\va{{\bm{a}}}

\def\vc{{\bm{c}}}

\def\ve{{\bm{e}}}

\def\vg{{\bm{g}}}

\def\vx{{\bm{x}}}
\def\vy{{\bm{y}}}
\def\vz{{\bm{z}}}

\def\eva{{a}}

\def\mA{{\bm{A}}}

\def\mH{{\bm{H}}}
\def\mI{{\bm{I}}}
\def\mJ{{\bm{J}}}

\def\mX{{\bm{X}}}

\def\mSigma{{\bm{\Sigma}}}

\DeclareMathAlphabet{\mathsfit}{\encodingdefault}{\sfdefault}{m}{sl}
\SetMathAlphabet{\mathsfit}{bold}{\encodingdefault}{\sfdefault}{bx}{n}
\newcommand{\tens}[1]{\bm{\mathsfit{#1}}}
\def\tA{{\tens{A}}}

\def\tX{{\tens{X}}}

\def\gG{{\mathcal{G}}}

\def\sA{{\mathbb{A}}}
\def\sB{{\mathbb{B}}}

\def\sS{{\mathbb{S}}}

\def\emA{{A}}

\newcommand{\etens}[1]{\mathsfit{#1}}

\def\etA{{\etens{A}}}

\newcommand{\E}{\mathbb{E}}

\newcommand{\R}{\mathbb{R}}

\newcommand{\KL}{D_{\mathrm{KL}}}
\newcommand{\Var}{\mathrm{Var}}

\newcommand{\Cov}{\mathrm{Cov}}

\newcommand{\normltwo}{L^2}
\newcommand{\normlp}{L^p}

\newcommand{\parents}{Pa} 

\newcommand{\vepsilon}{\bm{\epsilon}}

\usepackage{hyperref}
\usepackage{url}
\usepackage{graphicx}
\usepackage{booktabs}
\usepackage{twemojis}
\usepackage{colortbl}
\usepackage{tabularx}

\title{\texttwemoji{salt}\hspace{0.15em}Salt++: Context-Aligned Post-Training for Few-Step Streaming Multimodal Generation}

\author{%
  Xingtong Ge\textsuperscript{1,2}, 
  Yutong Wang\textsuperscript{3},
  Lunjie Zhu\textsuperscript{1},
  Haitao Lin\textsuperscript{4},
  Fangyu Lin\textsuperscript{1},\\
  \textbf{Yushi Huang\textsuperscript{1},}
  \textbf{Xin Zhang\textsuperscript{2},}
  \textbf{Yi Zhang\textsuperscript{2}\thanks{Project Lead}, }
  \textbf{Yu Liu\textsuperscript{2}, }
  \textbf{Jun Zhang\textsuperscript{1}\thanks{Corresponding Author}}  \\
  $^1$The Hong Kong University of Science and Technology,
  $^2$Vivix Group Limited \\
  $^3$The University of Sydney,
  $^4$Westlake University\\
  \texttt{xingtong.ge@gmail.com}
}

\iclrfinalcopy 
\begin{document}

\maketitle

\begin{abstract}
Few-step streaming audio--video generation requires both causal modeling and step distillation, yet standard training recipes face two context-related challenges. Teacher forcing pairs clean history with a noisy target, but supervises predictive contextual representations only indirectly through velocity prediction. Meanwhile, directly reusing bidirectional score models in causal Distribution Matching Distillation (DMD) creates a mismatch between generation and scoring contexts. We address these challenges with \textbf{Salt++}, a two-stage post-training framework comprising Causal Self-Flow (CSF) and context-aligned autoregressive DMD. CSF exploits contextual information asymmetry by varying the history while keeping the noisy target fixed: a noise-mixed-history student aligns its intermediate representations with those of a clean-history exponential-moving-average teacher. This self-supervised signal encourages the student to extract semantic information and improves cross-modal alignment. Context-aligned AR DMD shares the causal mask and prefix across generator sampling, fake-score training, and real-score evaluation to match generated and reference distributions under a block-conditional KL objective. With calibrated teacher guidance, it performs clean-prefix few-step distillation and then adapts to generated histories without switching objectives or requiring separate consistency distillation. At 480p, Salt++ improves visual and motion quality by 57\% and 45\% over OmniForcing on JavisBench under the same 4-step causal setting. A separate scale-wise post-training stage extends Salt++ to 4-step $1664\times960$ generation, outperforming bidirectional LTX-2 on six of seven reported metrics. Project page: \href{https://xingtongge.github.io/Saltpp}{https://xingtongge.github.io/Saltpp}.

\end{abstract}

\section{Introduction}
\label{sec:introduction}

Recent video generation models~\citep{yang2025cogvideox,kong2024hunyuanvideo,wan2025wan,gao2025seedance} can synthesize increasingly realistic scenes, and unified audio--video models~\citep{hacohen2026ltx2,chern2026speed,team2025kling,seedance2026seedance} further produce sound, speech, and visual content within a single diffusion process. Yet two properties put them at odds with real-time interaction. These models must generate an entire clip before displaying any output due to temporally bidirectional architectures, while modeling instantaneous velocity fields typically requires multi-step ODE integration for high-quality sampling, which is slow and expensive.

\begin{figure*}[t]
    \centering
    \includegraphics[width=0.92\textwidth]{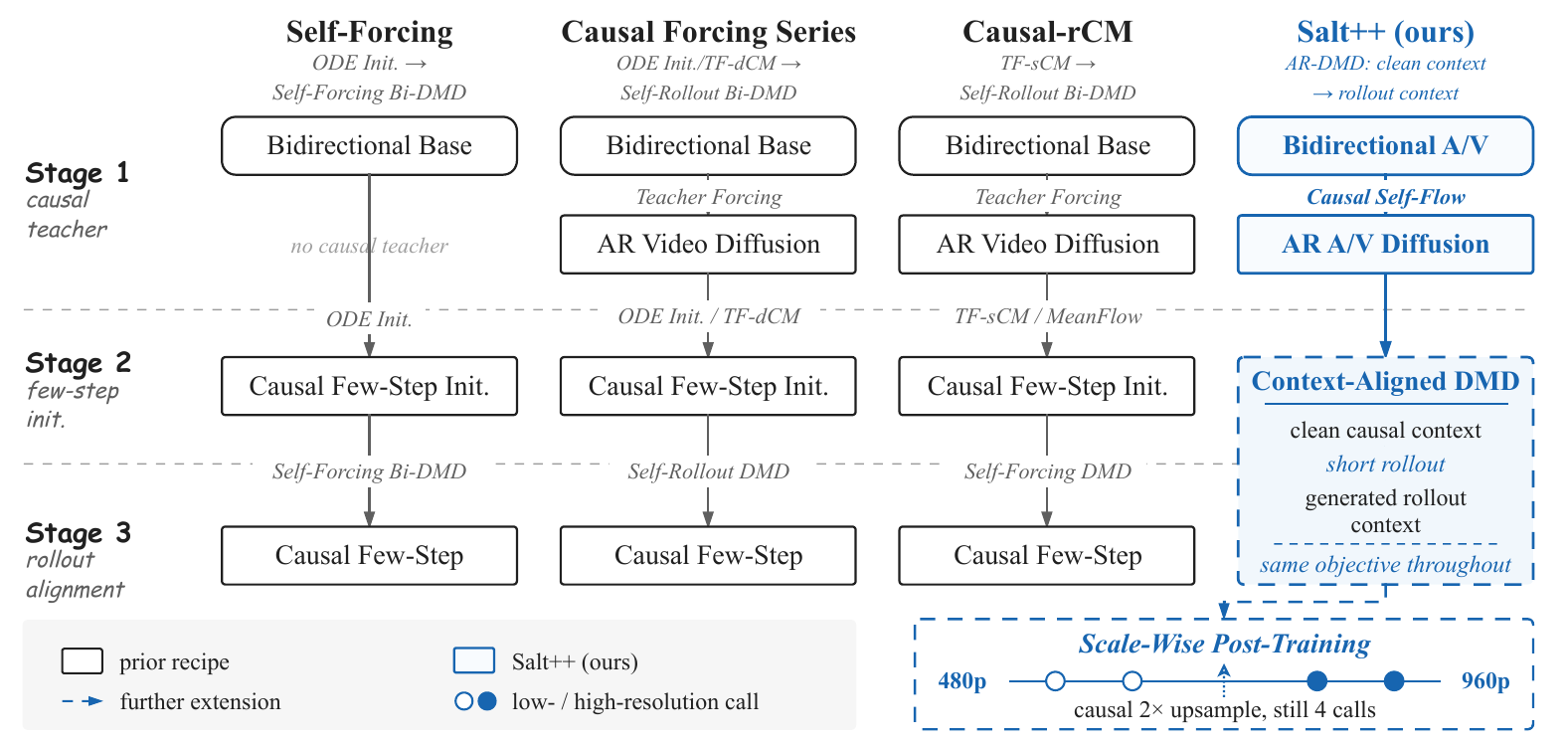}
    \vspace{-0.3cm}
    \caption{\textbf{Comparison with recent causal post-training recipes.}
    Prior recipes switch objectives between multiple stages and score the causal generator with bidirectional models (BI-DMD). Salt++ keeps one AR DMD objective throughout and only shifts its conditioning from clean context to generated rollout; a further scale-wise stage reaches $1664\times960$ with four generator calls.}
    \label{fig:saltpp_recipe}
    \vspace{-0.5cm}
\end{figure*}

Removing both costs asks for progress along two separate dimensions. The first is \emph{causal modeling}: autoregressive (AR) generation replaces temporally bidirectional attention with block-causal attention~\citep{yin2025causvid,huang2025selfforcing,zhu2026causal,ge2026salt}, so that each block is produced conditioned only on preceding ones and output can stream with bounded latency. The second is \emph{step distillation}: the sampling trajectory is compressed to a few evaluations while preserving the quality of the bidirectional teacher. However, prevailing recipes pursue the two dimensions through a long pipeline that changes objective at every stage---a causal teacher obtained by teacher forcing, a few-step generator initialized from it by ODE matching or consistency distillation, and a final refinement by distribution matching on the generator's own rollouts~\citep{lin2025aapt,zhao2026causalforcingpp,zheng2026causalrcm}; Fig.~\ref{fig:saltpp_recipe} lays these out.

These two dimensions raise distinct challenges concerning causal context. First, causal modeling requires extracting semantic information from past audio--video blocks that is predictive of the current block. Teacher forcing provides clean history alongside a noisy target, yet standard flow matching learns contextual representations only indirectly through velocity prediction. The challenge is to turn the information asymmetry into a direct learning signal for predictive contextual representations. Second, block-conditional Distribution Matching Distillation (DMD)~\citep{yin2024dmd,yin2024dmd2} requires generation and score estimation to share the same conditioning context. Evaluating score models with bidirectional context for blocks generated under a causal prefix introduces a \emph{score--context mismatch}, misaligning the score estimates with the intended block-conditional objective.

We address the first challenge with \textbf{Causal Self-Flow} (CSF). Motivated by semantic representation alignment~\citep{yu2025repa}, we adapt Self-Flow's asymmetric-view learning~\citep{chefer2026selfflow} to causal audio--video generation. A student observes a noise-mixed history, while an exponential-moving-average (EMA) teacher observes the corresponding clean history; both receive the same noisy target block. Alongside the native flow-matching objective, we align projected features from a shallow student layer with features from a deeper EMA-teacher layer, both extracted from the same noisy target block. By placing the input asymmetry entirely in the history, CSF encourages the student to recover clean-context features from partially corrupted history and learn contextual representations that support next-block prediction. Empirically, CSF improves audio--visual and audio--text alignment during causal teacher training (Fig.~\ref{fig:stage0_alignment_curves}).

We address the second challenge with \textbf{context-aligned AR DMD}. Generator sampling, fake-score training, and real-score evaluation share the same block-causal mask and audio--video prefix. The fake score therefore estimates the distribution induced by the causal generator under that prefix, while the real score represents the corresponding conditional data distribution, making resulting update an estimator of the gradient of a block-conditional KL objective. With aligned contexts and calibrated teacher guidance, DMD directly performs strong few-step distillation, replacing the separate ODE-matching or consistency-distillation stage used by prevailing recipes. We then adapt to generated histories while retaining the same objective, unifying clean-prefix distillation and on-policy adaptation.

Together these form the two-stage core of \textbf{Salt++}, yielding a 4-step causal audio--video generator. A separate scale-wise post-training stage distributes the same four generator evaluations across two spatial scales through a causal latent upsampler, extending the model to $1664\times960$ without additional generator calls.
In summary, our contributions are threefold:
\begin{itemize}
    \item We introduce Causal Self-Flow, turning asymmetric causal histories into a self-supervised representation-prediction task: a noise-mixed-history student learns from a clean-history teacher under the same noisy target, improving contextual audio--video representations.
    \item We identify score--context mismatch in block-conditional DMD and introduce context-aligned AR DMD. With calibrated teacher guidance, it unifies clean-prefix few-step distillation and on-policy adaptation within a single distillation stage.
    \item We obtain a 4-step causal audio--video generator that improves visual and motion quality by 57\% and 45\% over OmniForcing~\citep{su2026omniforcing}, and extend it to 4-step $1664\times960$ generation through scale-wise post-training, where it outperforms bidirectional LTX-2~\citep{hacohen2026ltx2} on six of seven reported JavisBench metrics.
\end{itemize}

\section{Related Work}
\label{sec:related_work}


\textbf{Audio--video generation and representation alignment.}
Joint audio--video diffusion models synthesize both modalities within a shared generative process. Existing systems explore hierarchical spatio-temporal priors, asymmetric or single-stream cross-modal interaction, and large-scale training~\citep{liu2025javisdit,low2025ovi,zhang2025uniavgen,team2026mova,chern2026speed,team2025kling,seedance2025seedance,seedance2026seedance}. We build on bidirectional LTX-2~\citep{hacohen2026ltx2} to obtain causal few-step generation. For representation learning, REPA aligns model features with an external visual encoder~\citep{yu2025repa}, while Self-Flow uses cleaner features from an EMA branch~\citep{chefer2026selfflow}. CSF adapts the latter principle to causal audio--video history, aligning a noise-mixed-history student with a clean-history EMA teacher under the same noisy target.

\textbf{Few-step distillation and autoregressive generation.}
Few-step generation commonly relies on trajectory or score based distillation~\citep{song2023consistency,luo2023latent,lin2026design,yin2024dmd,yin2024dmd2,pmlr-v267-lin25m,wang2026vdot}, with DMD extending to large scale image and video models~\citep{ge2026senseflow,ge2026salt}. Autoregressive diffusion models generate temporal blocks sequentially under causal conditioning~\citep{chen2024diffusionforcing,jin2025pyramidal,teng2025magi}. Recent methods combine causal generation with few-step distillation: CausVid distills bidirectional models into causal generators~\citep{yin2025causvid}, Self Forcing and AAPT adapt students on previously generated frames~\citep{huang2025selfforcing,lin2025aapt}, while Causal Forcing variants and Causal-rCM separate teacher-forced initialization from on-policy refinement~\citep{zhu2026causal,zhao2026causalforcingpp,zheng2026causalrcm}. Concurrent CMD similarly uses causal DMD~\citep{bandyopadhyay2026cmd}, but studies video-only distillation and moves directly to on-policy rollouts. We differ in three respects: we isolate the effect of score context in a controlled teacher-forced comparison, we show that the teacher guidance scale must be recalibrated for distribution matching rather than inherited from consistency distillation, and we obtain the AR teacher itself through a representation-alignment objective for joint audio--video generation.

\section{Method}
\label{sec:method}

\subsection{Problem Setup}
\label{sec:problem_setup}

\textbf{Causal audio--video generation.}
Let $\vy$ denote the text condition and $\vz_{1:K}$ a sequence of temporally aligned audio--video blocks, where $\vz_k=(\vz_k^v,\vz_k^a)$ groups the video and audio latents. Causal generation factorizes as
\begin{equation}
    p(\vz_{1:K}\mid\vy)
    =\prod\nolimits_{k=1}^{K}p(\vz_k\mid\vc_k),
    \qquad \vc_k=(\vy,\vz_{<k}).
    \label{eq:ar_factorization}
\end{equation}
A block-causal mask allows joint audio–video modeling within each block while restricting temporal context to preceding blocks. 
We denote $\vc_k^\star$ for the clean ground-truth context.


\textbf{Conditional flow matching.}
For a clean block $\vz_k$ and Gaussian noise $\vepsilon_k\sim\mathcal N(0,\mI)$, the linear flow path is~\citep{lipman2023flow,liu2022rectifiedflow}
\begin{equation}
    \vz_{k,t}=(1-t)\vz_k+t\vepsilon_k,
    \qquad t\in[0,1],
    \label{eq:flow_path}
\end{equation}
where larger $t$ means more noise. The flow model $F_\eta$ learns to predict velocity through
\begin{equation}
    \mathcal L_{\mathrm{FM}}
    =\mathbb E\!\left[
        \left\|F_\eta(\vz_{k,t},\vc_k,t)-(\vepsilon_k-\vz_k)\right\|_2^2
    \right].
    \label{eq:conditional_fm}
\end{equation}
Teacher forcing sets $\vc_k=\vc_k^\star$, conditioning the noisy current block on clean ground-truth history.

\textbf{Block-conditional distribution matching.}
Given a fixed $\vc_k$, a few-step generator $G_\theta$ induces $p_{\theta,k}(\cdot\mid\vc_k)$, with samples $\hat{\vz}_k^G$. Re-noising a generated block with Gaussian noise gives $\tilde{\vz}_{k,t}=(1-t)\hat{\vz}_k^G+t\vepsilon_k$, with $\tilde{\vz}_{k,t}\sim p_{\theta,k,t}(\cdot\mid\vc_k)$. Let $p_{\mathrm r,k,t}$ denote the conditional reference distribution at the same noise level; a real-score model approximates its score. The DMD objective is
\begin{equation}
    \mathcal L_{\mathrm{DMD},k}(\theta;\vc_k)
    =\mathbb E_t\!\left[
        \mathrm{KL}\!\left(
            p_{\theta,k,t}(\cdot\mid\vc_k)
            \,\Vert\,
            p_{\mathrm r,k,t}(\cdot\mid\vc_k)
        \right)
    \right].
    \label{eq:conditional_dmd}
\end{equation}
An online fake-score model learns the generator's conditional score~\citep{song2021sde} from generated samples. With exact conditional scores, the fake--real score difference supplies the score term in the gradient of Eq.~\ref{eq:conditional_dmd}~\citep{yin2024dmd}.

\begin{figure*}[t]
    \centering
    \includegraphics[width=\textwidth]{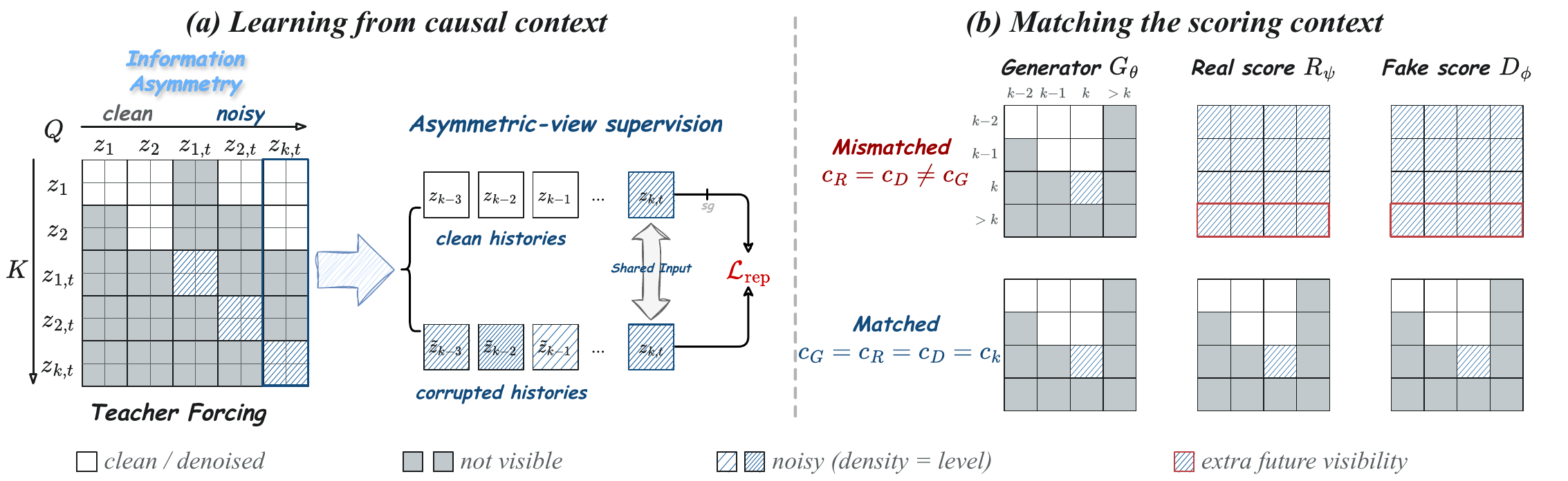}
    \vspace{-0.5cm}
    \caption{\textbf{Two roles of causal context in post-training.} \textbf{(a)} CSF exploits information asymmetry in causal histories for self-supervised representation learning. \textbf{(b)} Mismatched (top) and aligned (bottom) contexts across the generator, real score, and fake score.}
    \label{fig:context_challenges}
    \vspace{-0.5cm}
\end{figure*}

\textbf{Learning predictive contextual representations.}
For $k>1$, the model observes clean history alongside a noisy current block; 
the history provides temporal cues for denoising that block 
(Fig.~\ref{fig:context_challenges}(a)). Yet Eq.~\ref{eq:conditional_fm} supervises contextual representations only indirectly through velocity prediction. We seek an additional self-supervised signal that exploits this information asymmetry to learn predictive contextual representations (Sec.~\ref{sec:method_csf}).

\textbf{Score--context mismatch.}
Eq.~\ref{eq:conditional_dmd} compares distributions under the same $\vc_k$. Write $\vc_G$, $\vc_R$, and $\vc_D$ for the contexts used in generator sampling, real-score evaluation, and fake-score training. Additional future visibility or differently corrupted history changes the conditional information used for scoring (Fig.~\ref{fig:context_challenges}(b)). The challenge is to align generation and score estimation with the intended block-conditional objective, accounting for both history content and temporal visibility (Sec.~\ref{sec:method_context_dmd}).

\subsection{Causal Self-Flow}
\label{sec:method_csf}

Causal Self-Flow (CSF) turns history information asymmetry into a representation-learning task by pairing two causal views of the same denoising problem (Fig.~\ref{fig:context_challenges}(a)). An online student $F_\eta$ observes noise-mixed history, while its EMA teacher $F_{\bar\eta}$ observes clean history; both process the same noisy current block. We align their intermediate representations alongside the native flow-matching objective, encouraging the student to recover predictive contextual information from its mixed history.


\textbf{Clean and noise-mixed histories.}
For each target block $k$, we construct $\vz_{k,t_k}$ according to Eq.~\ref{eq:flow_path}. The student and EMA-teacher branches share the target block, Gaussian noise $\vepsilon_k$, and noise level $t_k$; their input asymmetry lies entirely in the history. For each preceding block $i<k$, we independently sample $\gamma_i\sim\mathcal{U}(\gamma_{\min},\gamma_{\max})$ and construct
\begin{equation}
    \tilde{\vz}_i^m
    =
    (1-\gamma_i)\vz_i^m+\gamma_i\vepsilon_i^m,
    \qquad
    \vepsilon_i^m\sim\mathcal{N}(0,\mI),
    \quad m\in\{v,a\}.
    \label{eq:csf_context_noise}
\end{equation}
The temporally aligned video and audio portions share the same $\gamma_i$, while their Gaussian noise is sampled independently. 
The student receives the noise-mixed context $\tilde{\vc}_k=(\vy,\tilde{\vz}_{<k})$, whereas the EMA teacher receives the clean context $\vc_k^\star$. Thus, history corruption changes the available contextual information without changing the current block's denoising task.

\textbf{Cross-view, cross-depth representation alignment.}
Let $H_{\eta,m}^{\ell}$ and $H_{\bar\eta,m}^{\ell}$ denote the representations of modality $m$ at layer $\ell$ of the student and EMA teacher, respectively. A two-layer projection head $P_m$ maps the shallower student representation at layer $\ell_s$ to the deeper EMA-teacher representation at layer $\ell_d$:
\begin{equation}
    \mathcal{L}_{\mathrm{rep}}^m
    =
    \mathbb{E}_{k>1}\bigl[
        1-\cos\bigl(
            P_m\bigl(H_{\eta,m}^{\ell_s}(\vz_{k,t_k},\tilde{\vc}_k,t_k)\bigr),\,
            \operatorname{sg}\bigl[H_{\bar\eta,m}^{\ell_d}(\vz_{k,t_k},\vc_k^\star,t_k)\bigr]
        \bigr)
    \bigr],
    \label{eq:csf_rep}
\end{equation}
where $\operatorname{sg}$ denotes stop-gradient. We apply representation alignment only to blocks with preceding history. The clean-history teacher provides a contextual representation target for the mixed-history student, rather than a target for matching final denoising predictions.
The student also minimizes the video and audio flow-matching losses under $\tilde{\vc}_k$. The combined objective is
\begin{equation}
    \mathcal{L}_{\mathrm{CSF}}
    =
    \mathcal{L}_{\mathrm{FM}}^v
    +\lambda_a\mathcal{L}_{\mathrm{FM}}^a
    +\lambda_{\mathrm{rep}}
    \left(
        \mathcal{L}_{\mathrm{rep}}^v
        +\lambda_a\mathcal{L}_{\mathrm{rep}}^a
    \right),
    \label{eq:csf_total}
\end{equation}
where $\lambda_a$ balances the modalities and $\lambda_{\mathrm{rep}}$ weights representation alignment. We alternate CSF updates with standard clean-history teacher-forcing updates with equal probability. After training, the student serves as the AR teacher for subsequent distillation; the EMA branch and projection heads are discarded, introducing no inference-time overhead. Fig.~\ref{fig:stage0_alignment_curves} shows that CSF improves audio--visual and audio--text alignment relative to standard teacher forcing in causal teacher training.

\subsection{Context-Aligned AR DMD for Few-Step Causal Generation}
\label{sec:method_context_dmd}

\begin{figure*}[t]
    \centering
    \includegraphics[width=0.95\textwidth]{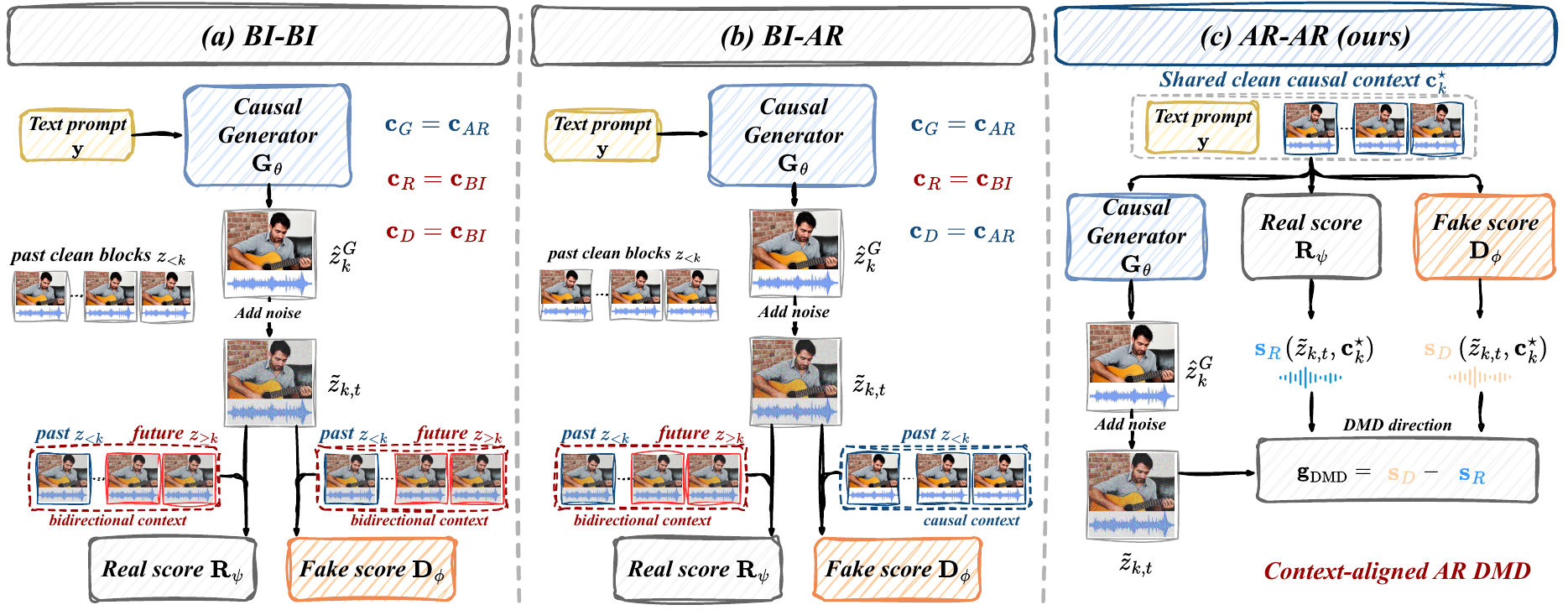}
    \vspace{-0.2cm}
    \caption{\textbf{Score--context configurations in causal DMD.}
    The generator always samples under a causal prefix, while the two scores may use bidirectional (BI) or causal (AR) visibility. Only AR--AR scores each block under the aligned context, estimating the gradient of Eq.~\ref{eq:conditional_dmd}.}
    \label{fig:stage2_context_aligned_dmd}
    \vspace{-0.5cm}
\end{figure*}

Building on the causal model trained by CSF, we perform few-step distillation with a generator $G_\theta$, a frozen real-score model $R_\psi$, and an online fake-score model $D_\phi$. Context-aligned AR DMD addresses score--context mismatch by placing sample generation, fake-score training, and real-score evaluation under the same causal context. We first instantiate the block-conditional objective in Eq.~\ref{eq:conditional_dmd} under clean prefixes, then extend it to generated histories without switching distillation objectives.

\textbf{Clean-prefix sampling and fake-score training.}
Under teacher forcing, the generator performs few-step sampling conditioned on $\vc_k^\star$, producing clean blocks $\hat{\vz}_k^G\sim p_{\theta,k}(\cdot\mid\vc_k^\star)$. The clean prefix is held fixed throughout sampling. To estimate the score of this generator-induced distribution, the fake model must be trained on generated blocks paired with the contexts under which they were sampled.
We form the perturbation $\tilde{\vz}_{k,t}$ of Sec.~\ref{sec:problem_setup} from a fresh generated block, with $t$ and $\vepsilon_k$ sampled independently of the generator's sampling procedure. Using the flow-velocity parameterization, we train the fake model with
\begin{equation}
    \mathcal{L}_{\mathrm{fake}}
    =
    \mathbb{E}\!\left[
        \left\|
            D_\phi\!\left(
                \operatorname{sg}(\tilde{\vz}_{k,t}),
                \vc_k^\star,t
            \right)
            -
            \left(
                \vepsilon_k-\operatorname{sg}(\hat{\vz}_k^G)
            \right)
        \right\|_2^2
    \right].
    \label{eq:conditional_fake_matching}
\end{equation}
The generated block is detached during fake-score training, and the fake model receives the same block packing, causal mask, and clean prefix as the generator. This objective learns the conditional flow field corresponding to $p_{\theta,k,t}(\cdot\mid\vc_k^\star)$, from which its score can be obtained. Changing the conditioning would instead change the conditional distribution being estimated.

\textbf{Context-aligned real-score evaluation.}
With the fake score tied to the generator distribution, the real score specifies the reference to be distilled. The CSF-trained AR teacher provides a reference for next-block generation under a causal prefix. Using this teacher allows few-step distillation to transfer the causal generation capability learned in Sec.~\ref{sec:method_csf}, with the teacher and generator evaluated on the same next-block prediction task.

Fig.~\ref{fig:stage2_context_aligned_dmd} compares three score configurations, where the first and second entries denote the visibility patterns of the real and fake scores, respectively. In BI--BI, a common choice in prior work~\citep{zhu2026causal,zheng2026causalrcm,su2026omniforcing}, both scores use bidirectional visibility that differs from the causal context used to generate each block. BI--AR aligns the fake score with the generator while retaining a bidirectional real model. AR--AR instead uses the causal teacher as the reference, placing sample generation and both score models under the same block packing, causal mask, and prefix. Our controlled comparison shows that AR--AR achieves the best overall performance among these configurations (Sec.~\ref{sec:stage2_results}), validating the effectiveness of our context-aligned AR DMD formulation for few-step causal distillation.

\textbf{Generator update and teacher guidance.}
Under AR--AR, the real and fake models evaluate the same perturbed block $\tilde{\vz}_{k,t}$ at the same noise level $t$ and clean prefix $\vc_k^\star$. With exact conditional scores, their difference supplies the score-difference term in the gradient of Eq.~\ref{eq:conditional_dmd}. In practice, we use their clean predictions $\hat{\vz}_k^{R}$ and $\hat{\vz}_k^{D}$ to form a normalized direction for each modality $m\in\{v,a\}$, which is injected through a stop-gradient surrogate:
\begin{equation}
    \vg_k^m =
    \tfrac{\hat{\vz}_k^{D,m}-\hat{\vz}_k^{R,m}}
    {\operatorname{mean}\lvert\hat{\vz}_k^{G,m}-\hat{\vz}_k^{R,m}\rvert+\epsilon},
    \qquad
    \mathcal{L}_G^m =
    \tfrac{1}{2}\bigl\|\hat{\vz}_k^{G,m}
    -\operatorname{sg}(\hat{\vz}_k^{G,m}-\vg_k^m)\bigr\|_2^2,
    \label{eq:dmd_surrogate}
\end{equation}
where $\mathcal{L}_G=\mathcal{L}_G^v+\lambda_a\mathcal{L}_G^a$ combines the two modalities.
Besides, teacher CFG shapes the real-score target.
We further calibrate it for DMD, independently sampling video and audio guidance from a lower range at each update rather than inheriting the fixed high guidance used by consistency distillation (Sec.~\ref{sec:stage2_results}). This clean-prefix procedure performs few-step distillation directly through DMD, without a separate consistency-distillation stage.

\textbf{On-policy context adaptation.}
Clean-prefix DMD performs few-step distillation under ground-truth histories, whereas streaming inference conditions on the generator's own predictions. To reduce this train--test context gap, we continue training on autoregressively generated histories, replacing $\vc_k^\star$ with $\hat{\vc}_k=(\vy,\hat{\vz}_{<k}^G)$ in Eq.~\ref{eq:conditional_dmd}. Generator sampling, fake-score training, and real-score evaluation remain conditioned on the same context. This preserves the block-conditional objective while adapting the conditioning distribution to the generated histories encountered during inference.

\begin{figure}[t]
    \centering
    \includegraphics[width=0.9\linewidth]{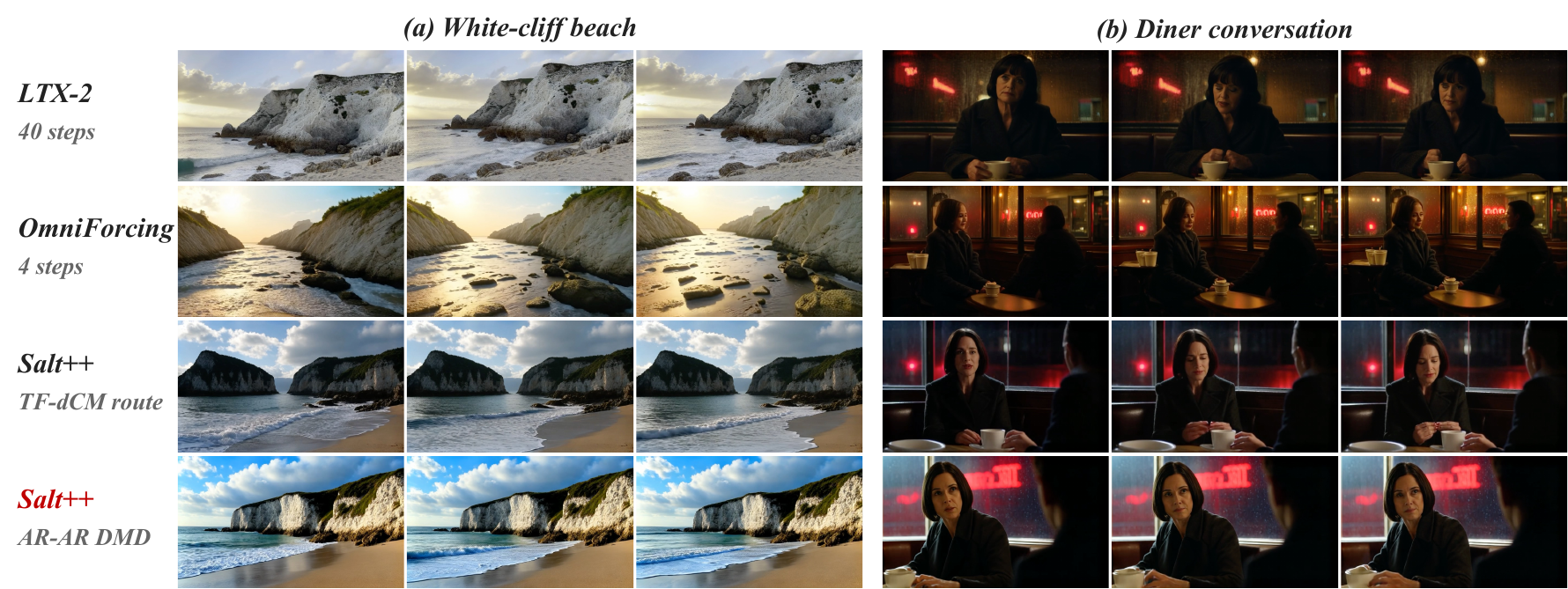}
    \vspace{-0.3cm}
    \caption{\textbf{Qualitative comparison of 480p generation.}
    Coast and diner examples compare LTX-2 Base (40 steps), OmniForcing, and
    both Salt++ routes (4 steps).
    The AR--AR DMD route retains fine scene and facial detail.
    Complete prompts and more comparisons are provided in Appendix~\ref{sec:visual_examples}.}
    \label{fig:main_coast_diner}
    \vspace{-0.3cm}
\end{figure}

\subsection{Scale-Wise High-Resolution Post-Training}
\label{sec:method_stage4}

As a further extension, we build on the 4-step streaming generator obtained in Sec.~\ref{sec:method_context_dmd} to enable high-resolution generation. Specifically, we train the generator itself to perform both low-resolution (LR) generation and high-resolution (HR) refinement, allocating two denoising steps to each scale.

\textbf{Chunk-wise cross-scale generation.}
For each temporal block, the generator first performs two LR evaluations to produce a clean audio--video prediction. A causal latent upsampler $U_\omega$ then spatially upsamples the video latent using only the current block and a bounded history window, while the audio latent retains its resolution. Both modalities are re-noised to initialize two HR evaluations, after which the completed block is output before proceeding to the next block. During both training rollouts and inference, the generator shares parameters across scales but maintains separate LR and HR KV caches, each containing the corresponding audio--video history. The scales communicate through the latent transition. We initialize $U_\omega$ by distilling the released LTX-2 latent upsampler into this windowed causal form. This block-wise procedure preserves causal streaming while producing $1664\times960$ output with four generator evaluations per block.

\textbf{Scale-wise distribution matching.}
The CSF teacher is post-trained at 480p, so we use a high-resolution-capable bidirectional teacher to supervise this extension. The generator remains causal and is trained on its own rollouts. A frozen real-score model and an online fake-score model, shared across scales, evaluate the completed LR and HR rollouts with bidirectional visibility. Unlike the block-conditional AR--AR matching in Sec.~\ref{sec:method_context_dmd}, this stage matches distributions at the scale-rollout level. The fake model learns from fresh generator rollouts, and we apply the DMD surrogate in Eq.~\ref{eq:dmd_surrogate} to the joint audio--video output at each scale:
\begin{equation}
    \mathcal{L}_{\mathrm{scale}} = \sum\nolimits_{s\in\{L,H\}} \mathcal{L}_G(\hat{\vz}_{1:K}^{s}),
    \label{eq:stage4_scale_loss}
\end{equation}
where $\hat{\vz}_{1:K}^{s}$ denotes the generated rollout at scale $s$. Both terms update the shared generator, while the HR term can additionally update the causal upsampler. 
Details are provided in Appendix~\ref{app:high_resolution}.

\begin{table*}[t]
    \centering
    \caption{\textbf{Main results on JavisBench-mini with official prompts.}
    (a,b) JavisBench at 480p/960p; (c) VBench at 480p.
    At 480p, bold/underline indicate the best/second-best 4-step results.
    At 960p, bold marks the best result across all methods}
    \label{tab:javisbench_main}
    \small
    \setlength{\tabcolsep}{2.0pt}
    \begin{tabular}{lccrrrrrrr}
        \toprule
        Model & Causal & Steps & VQ $\uparrow$ & MQ $\uparrow$ & AQ $\uparrow$ & CLIP $\uparrow$ & IB-AV $\uparrow$ & Javis $\uparrow$ & DeSync $\downarrow$ \\
        \midrule
        \multicolumn{10}{l}{\textit{(a) 480p generation}} \\
        LTX-2 Base~\citep{hacohen2026ltx2} & No & 40 & 1.884 & 0.566 & 4.986 & 0.311 & 0.239 & 0.200 & 0.608 \\
        AR teacher (CSF) & Yes & 40 & 2.304 & 0.857 & 4.565 & 0.316 & 0.202 & 0.162 & 0.746 \\
        OmniForcing~\citep{su2026omniforcing} & Yes & 4 & 1.807 & 0.699 & 4.718 & 0.303 & 0.163 & 0.124 & \underline{0.745} \\
        Salt++ (TF-dCM route) & Yes & 4 & \underline{2.013} & \underline{0.826} & \underline{4.976} & \underline{0.313} & \textbf{0.229} & \textbf{0.185} & \textbf{0.710} \\
        \rowcolor{rowgray}\textbf{Salt++ (AR--AR DMD route)} & Yes & 4 & \textbf{2.838} & \textbf{1.010} & \textbf{4.991} & \textbf{0.316} & \underline{0.184} & \underline{0.146} & 0.759 \\
        \midrule
        \multicolumn{10}{l}{\textit{(b) 960p generation}} \\
        LTX-2 Base~\citep{hacohen2026ltx2} & No & $40+3$ & 2.231 & 0.607 & 4.867 & 0.311 & 0.169 & 0.145 & \textbf{0.658} \\
        OmniForcing~\citep{su2026omniforcing} & Yes & 4 & 2.298 & 0.832 & 4.721 & 0.293 & 0.165 & 0.128 & 0.755 \\
        \rowcolor{rowgray}\textbf{Salt++} & Yes & 4 & \textbf{2.730} & \textbf{0.957} & \textbf{5.113} & \textbf{0.318} & \textbf{0.193} & \textbf{0.157} & 0.768 \\
        \bottomrule
    \end{tabular}

    \vspace{1pt}

    \begin{tabularx}{0.985\textwidth}{>{\raggedright\arraybackslash}Xccrrrr}
        \toprule
        \multicolumn{7}{l}{\textit{(c) 480p VBench video quality and within-clip consistency}} \\
        \addlinespace[1pt]
        Model & Causal & Steps & Aesthetic $\uparrow$ & Imaging $\uparrow$ & Subject $\uparrow$ & Background $\uparrow$ \\
        \midrule
        LTX-2 Base~\citep{hacohen2026ltx2} & No & 40 & 53.89 & 67.06 & 95.94 & 95.63 \\
        OmniForcing~\citep{su2026omniforcing} & Yes & 4 & \textbf{56.74} & \underline{68.41} & \underline{96.62} & 95.32 \\
        Salt++ (TF-dCM route) & Yes & 4 & 53.65 & 62.87 & 96.58 & \underline{96.05} \\
        \rowcolor{rowgray}\textbf{Salt++ (AR--AR DMD route)} & Yes & 4
            & \underline{55.43} & \textbf{69.98} & \textbf{97.01} & \textbf{96.24} \\
        \bottomrule
    \end{tabularx}
    \vspace{-0.5cm}
\end{table*}
\begin{table*}[t]
    \centering
    \caption{\textbf{Quantitative results on the 1,000 LTX-2-enhanced JavisBench prompts at 480p.}}
    \label{tab:javisbench_rewrite}
    \small
    \setlength{\tabcolsep}{2.3pt}
    \begin{tabular}{lccrrrrrrr}
        \toprule
        Model & Causal & Steps & VQ $\uparrow$ & MQ $\uparrow$ & AQ $\uparrow$ & IB-AV $\uparrow$ & AVH $\uparrow$ & Javis $\uparrow$ & DeSync $\downarrow$ \\
        \midrule
        LTX-2 Base~\citep{hacohen2026ltx2} & No & 40 & 1.981 & 0.640 & 4.959 & 0.256 & 0.245 & 0.211 & 0.573 \\
        OmniForcing~\citep{su2026omniforcing} & Yes & 4 & 1.774 & 0.607 & 4.613 & 0.158 & 0.152 & 0.121 & 0.734 \\
        Salt++ (TF-dCM route) & Yes & 4 & \underline{2.074} & \underline{0.760} & \underline{4.838} & \textbf{0.245} & \textbf{0.233} & \textbf{0.193} & \underline{0.704} \\
        \rowcolor{rowgray}Salt++ (AR--AR DMD route) & Yes & 4 & \textbf{2.820} & \textbf{0.999} & \textbf{4.997} & \underline{0.192} & \underline{0.186} & \underline{0.151} & \textbf{0.696} \\
        \bottomrule
    \end{tabular}
    \vspace{-0.5cm}
\end{table*}

\section{Experiments}
\label{sec:experiments}

\subsection{Implementation Details}
\label{sec:implementation}

\textbf{Models and baselines.}
We build Salt++ on LTX-2~\citep{hacohen2026ltx2}, using LTX-2 Base as a multi-step reference and OmniForcing~\citep{su2026omniforcing} as the 4-step causal baseline. We also report the CSF AR teacher before step distillation and implement a TF-dCM comparison route following Causal Forcing++ and Causal-rCM~\citep{zhao2026causalforcingpp,zheng2026causalrcm}. Our AR--AR DMD route instead performs clean-prefix distillation and on-policy adaptation with the same context-aligned objective. Both complete routes use 4-step causal inference with CFG $=1$.

\textbf{Evaluation protocol.}
We evaluate on the 1,000 JavisBench-mini prompts~\citep{liu2025javisdit} and a fixed set of their LTX-2 prompt-enhancer (PE) rewrites, shared across models. At 480p, videos contain 121 frames at $832\times480$ and 24 FPS. We report JavisBench quality, semantic alignment, and synchronization metrics, alongside four VBench ~\cite{huang2024vbench} quality and consistency metrics. High-resolution evaluation uses $1664\times960$ output. Full sampling settings, metric definitions, and training configurations are provided in Appendix~\ref{app:implementation}.




\begin{figure*}[t]
    \centering
    \includegraphics[width=0.95\textwidth]{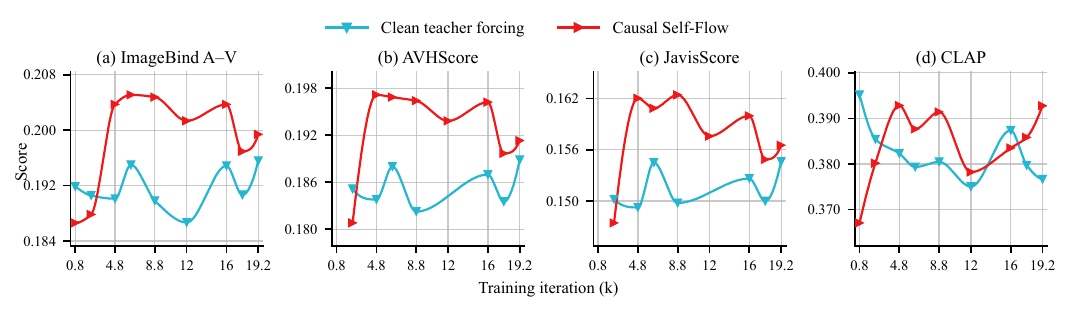}
    \vspace{-0.5cm}
    \caption{\textbf{Cross-modal alignment during AR teacher training.}
    Causal Self-Flow versus clean teacher forcing over 19.2k iterations, evaluated with 40-step inference on the 1,000 JavisBench prompts. Panels (a--c) report audio--visual agreement and panel (d) reports audio--text agreement.}
    \label{fig:stage0_alignment_curves}
    \vspace{-0.35cm}
\end{figure*}

\subsection{Main Results}
\label{sec:main_results}

\textbf{4-step generation at 480p.}
Across JavisBench-mini and VBench, our AR--AR DMD route outperforms the TF-dCM route on eight of eleven metrics, including all four VBench metrics (Tab.~\ref{tab:javisbench_main}(a,c)). It also achieves the best scores on seven metrics among all 4-step methods, improving visual and motion quality over OmniForcing by 57.1\% and 44.5\%, respectively. The TF-dCM route retains advantages in IB-AV, JavisScore, and DeSync, while OmniForcing leads in aesthetics. Fig.~\ref{fig:main_coast_diner} illustrates the visual-quality advantage of AR--AR DMD, with clearer scene textures and facial details.

\textbf{Generation with rewritten prompts.}
\label{sec:rewrite_results}
Tab.~\ref{tab:javisbench_rewrite} evaluates all models on the same 1,000 LTX-2-enhanced prompts. AR--AR DMD continues to outperform TF-dCM route on a majority of metrics, leading on VQ, MQ, AQ, and DeSync, while TF-dCM route retains higher audio--visual semantic alignment scores. Compared with OmniForcing, AR--AR DMD improves all seven reported metrics, with VQ and MQ increasing by 59.0\% and 64.6\%, respectively. These results show that its advantages extend from official prompts to richer rewritten descriptions.

\begin{figure}[t]
    \centering
    \includegraphics[width=0.9\linewidth]{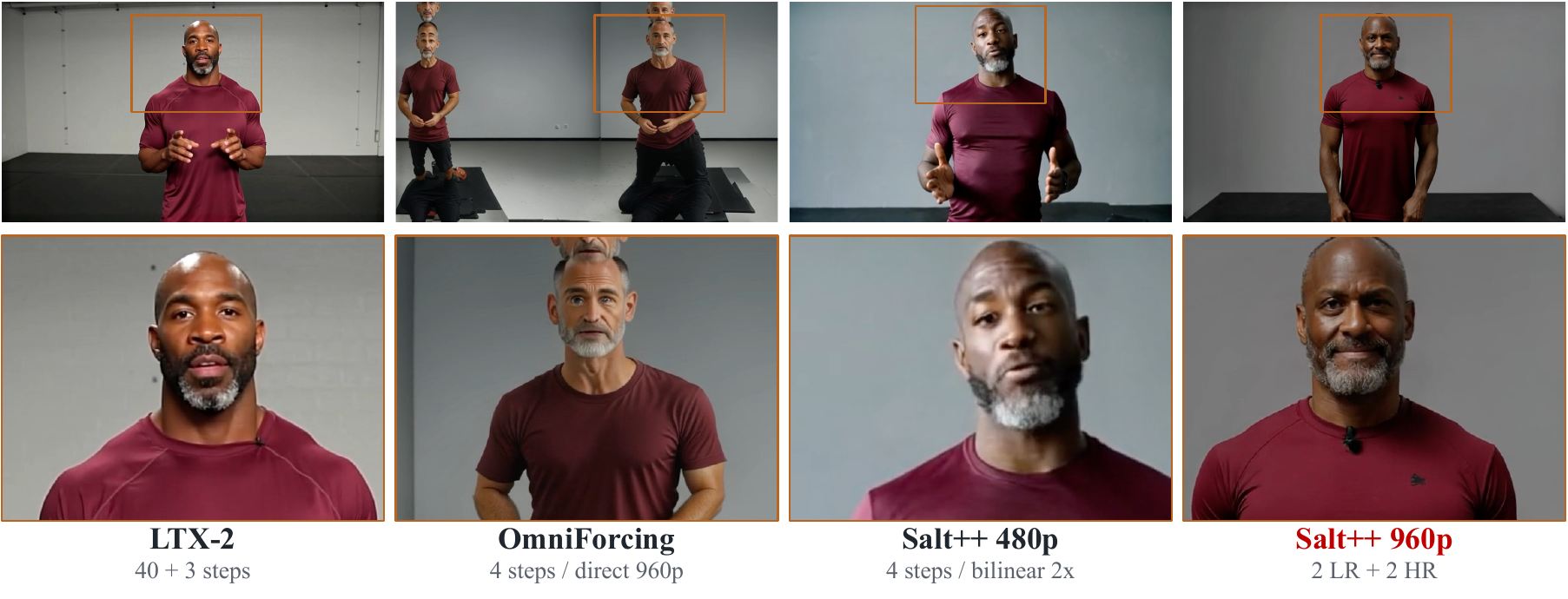}
    \vspace{-0.5cm}
    \caption{\textbf{960p generation: Fitness coach.}
    Columns compare LTX-2 (40+3 steps), OmniForcing extrapolated to
    960p, upsampled Salt++ 480p, and native Salt++ 960p. Full frames and detail windows show the finer facial detail of native 960p generation. More examples appear in
    Appendix~\ref{sec:hr_visuals}.}
    \vspace{-0.5cm}
    \label{fig:hr_fitness}
\end{figure}

\textbf{High-resolution extension.}
Scale-wise post-training extends Salt++ to $1664\times960$ generation with four generator evaluations per block. It outperforms the $40+3$-step LTX-2 reference on six of seven metrics in Tab.~\ref{tab:javisbench_main}(b), with VQ/MQ reaching 2.730/0.957 versus 2.231/0.607; DeSync remains the exception. The table also reports OmniForcing directly extrapolated to 960p, rather than a resolution-matched trained baseline. Fig.~\ref{fig:hr_fitness} shows finer facial detail in native 960p output than in the bilinearly upsampled 480p comparison. More qualitative results and full prompts are provided in Appendix~\ref{sec:visual_examples}.


\subsection{Ablation Studies and Analysis}
\label{sec:ablation_results}

\textbf{Causal Self-Flow.}
\label{sec:csf_results}
Fig.~\ref{fig:stage0_alignment_curves} compares CSF with standard teacher forcing using 40-step inference on the same 1,000 JavisBench prompts. After the initial optimization phase, CSF consistently improves IB-AV, AVHScore, and JavisScore, with CLAP gains emerging later in training. 
These improvements occur before few-step distillation, demonstrating stronger audio--visual and audio--text alignment in the causal teacher and supporting the effectiveness of CSF for contextual representation learning.

\begin{figure}[t]
    \centering
    \includegraphics[width=0.95\linewidth]{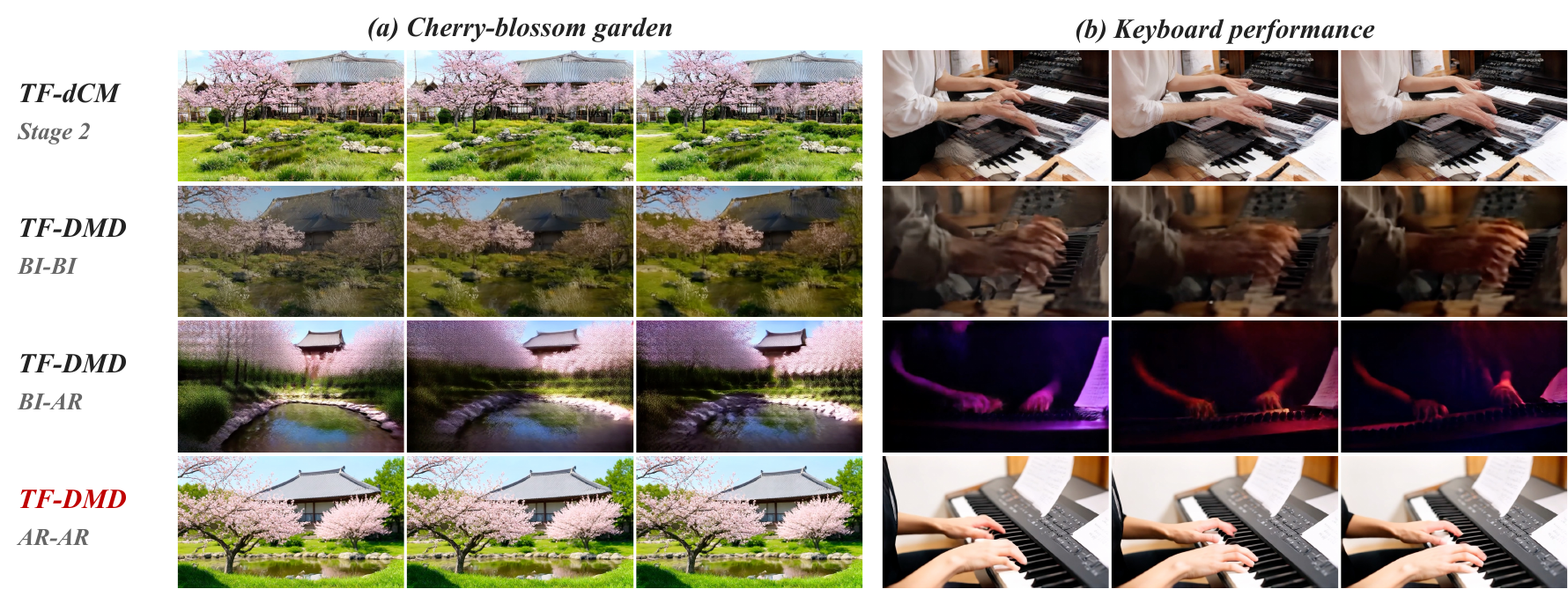}
    \vspace{-0.3cm}
    \caption{\textbf{Qualitative results of few-step initialization.}
    Rows compare TF-dCM with BI--BI, BI--AR, and AR--AR TF-DMD. AR--AR retains significantly better scene structure and details.}
    \label{fig:stage2_qualitative}
    \vspace{-0.5cm}
\end{figure}

\textbf{Score context and distillation objective.}
\label{sec:stage2_results}
Tab.~\ref{tab:stage2_context_ablation}(a) compares score configurations and distillation objectives under teacher forcing, before on-policy adaptation. Among the DMD configurations, aligning only the fake score raises VQ/MQ from 0.837/0.140 in BI--BI to 1.131/0.157 in BI--AR. The fully aligned AR--AR configuration reaches 3.147/1.351 and achieves the best scores on five of seven metrics; Appendix~\ref{app:metrics} discusses why DeSync favors the near-static BI--AR outputs. These results support aligning score models with the generator's causal context for effective block-conditional distillation. Besides, compared with TF-dCM, AR--AR TF-DMD improves all seven reported metrics, with VQ/MQ increasing from 1.916/0.611 to 3.147/1.351. This establishes AR DMD as an effective primary few-step distillation objective, without requiring a separate consistency-distillation stage. Fig.~\ref{fig:stage2_qualitative} illustrates the corresponding significant improvements in scene structure and detail.

\textbf{Teacher-guidance calibration.}
\label{sec:guidance_results}
Tab.~\ref{tab:stage2_context_ablation}(b) compares fixed video/audio teacher guidance of 4.0 with independent sampling from $\mathcal{U}(1.0,3.5)$, using the same AR--AR configuration, training iteration, and inference CFG. This combined adjustment of guidance range and randomization improves VQ by 91.5\% on official prompts and 82.9\% on rewritten prompts, while MQ more than doubles in both settings. Appendix~\ref{app:guidance_provenance} shows the corresponding qualitative difference in tonal and texture detail. These results highlight the importance of calibrating teacher guidance for DMD rather than directly transferring the consistency-distillation setting.

\textbf{On-policy context adaptation.}
\label{sec:onpolicy_results}
Following clean-prefix distillation, on-policy training adapts the generator to its own histories. Qualitative inspection shows reduced overexposure in outputs after adaptation. The complete-route results in Sec.~\ref{sec:main_results} report performance after this continuation, which retains the context-aligned DMD objective while changing the source of the causal history.

\begin{table*}[t]
    \centering
    \caption{\textbf{Few-step initialization ablations at 480p.}
    (a) Score contexts and objectives on official prompts.
    (b) Teacher guidance under aligned AR--AR scores, on both prompt sets.
    Bold/underline mark best/second-best results in (a); bold marks the better result in (b).}
    \vspace{1pt}
    \label{tab:stage2_context_ablation}
    \small
    \setlength{\aboverulesep}{0.15ex}\setlength{\belowrulesep}{0.25ex}
    \renewcommand{\arraystretch}{0.96}
    \begin{tabularx}{0.985\textwidth}{>{\raggedright\arraybackslash}Xcrrrrrrr}
        \toprule
        \textit{(a)} Objective & Scores & VQ $\uparrow$ & MQ $\uparrow$ & AQ $\uparrow$ & CLIP $\uparrow$ & IB-TV $\uparrow$ & IB-TA $\uparrow$ & DeSync $\downarrow$ \\
        \midrule
        TF-dCM & -- & \underline{1.916} & \underline{0.611} & 4.686 & \underline{0.307} & 0.264 & 0.145 & 0.766 \\
        TF-DMD & BI/BI & 0.837 & 0.140 & \textbf{4.796} & 0.301 & \underline{0.266} & 0.144 & 0.811 \\
        TF-DMD & BI/AR & 1.131 & 0.157 & 4.571 & 0.287 & 0.255 & \underline{0.149} & \textbf{0.531} \\
        \rowcolor{rowgray}\textbf{TF-DMD} & \textbf{AR/AR} & \textbf{3.147} & \textbf{1.351} & \underline{4.750} & \textbf{0.317} & \textbf{0.271} & \textbf{0.153} & \underline{0.726} \\
        \bottomrule
    \end{tabularx}
    
    \vspace{2pt}
    
    \begin{tabularx}{0.985\textwidth}{>{\raggedright\arraybackslash}Xrrrrrrr}
        \toprule
        \textit{(b)} Teacher guidance & \multicolumn{2}{c}{Official} & \multicolumn{2}{c}{Rewrite} & \multicolumn{3}{c}{Official diagnostics} \\
        \cmidrule(lr){2-3}\cmidrule(lr){4-5}\cmidrule(lr){6-8}
        & VQ $\uparrow$ & MQ $\uparrow$ & VQ $\uparrow$ & MQ $\uparrow$ & AQ $\uparrow$ & CLIP $\uparrow$ & DeSync $\downarrow$ \\
        \midrule
        Fixed $4.0$ & 1.643 & 0.572 & 1.712 & 0.598 & \textbf{4.934} & 0.315 & 0.843 \\
        \rowcolor{rowgray}\textbf{Randomized $\mathcal{U}(1.0,3.5)$} & \textbf{3.147} & \textbf{1.351} & \textbf{3.132} & \textbf{1.261} & 4.750 & \textbf{0.317} & \textbf{0.726} \\
        \bottomrule
    \end{tabularx}
    \vspace{-0.5cm}
\end{table*}

\section{Conclusion}
\label{sec:conclusion}
We presented Salt++, a context-aligned post-training framework for few-step streaming audio--video generation. Causal Self-Flow strengthens contextual learning in the AR teacher, while context-aligned AR DMD conditions the generator and both score models on the same causal prefix and keeps that objective as the prefix changes from ground truth to generated rollouts. Once the conditional paths agree, distribution matching is well posed enough to serve as the primary few-step objective, removing the separate consistency-distillation stage of prevailing recipes. The resulting 4-step generator improves visual and motion quality by 57\% and 45\% over OmniForcing at 480p, and scale-wise post-training reaches $1664\times960$ within the same budget.\label{sec:bodyend}

\clearpage
\bibliography{iclr2027_conference}
\bibliographystyle{iclr2027_conference}

\clearpage
\appendix
\begin{center}
  {\Large\bfseries Appendix\par}
\end{center}
\phantomsection
\label{app:contents}
\vspace{0.65em}

\begingroup
\color{black}
\hypersetup{hidelinks}
\setlength{\fboxsep}{8pt}
\setlength{\fboxrule}{0.4pt}
\noindent\fcolorbox{black}{white}{%
\begin{minipage}{\dimexpr\linewidth-2\fboxsep-2\fboxrule\relax}
\raggedright
{\normalsize Contents\par}
\vspace{0.45em}
\small
\setlength{\parskip}{0.12em}
\newcommand{\saltappsectionentry}[2]{%
  \noindent{\hyperref[#1]{%
    \makebox[2.2em][l]{\textbf{\ref*{#1}}}\textbf{#2}}}%
  \hfill\hyperref[#1]{\textbf{\pageref*{#1}}}\par
}
\newcommand{\saltappsubsectionentry}[2]{%
  {\footnotesize\noindent\hspace*{1.2em}%
  {\hyperref[#1]{%
    \makebox[3.2em][l]{\ref*{#1}}#2}}%
  \nobreak\leaders\hbox to 0.55em{\hss.\hss}\hfill\nobreak
  \hyperref[#1]{\pageref*{#1}}\par}%
}

{\footnotesize\itshape Main text\par}
\vspace{0.15em}
\saltappsectionentry{sec:introduction}{Introduction}
\saltappsectionentry{sec:related_work}{Related Work}
\saltappsectionentry{sec:method}{Method}
\saltappsubsectionentry{sec:problem_setup}{Problem Setup}
\saltappsubsectionentry{sec:method_csf}{Causal Self-Flow}
\saltappsubsectionentry{sec:method_context_dmd}{Context-Aligned AR DMD for Few-Step Causal Generation}
\saltappsubsectionentry{sec:method_stage4}{Scale-Wise High-Resolution Post-Training}
\saltappsectionentry{sec:experiments}{Experiments}
\saltappsubsectionentry{sec:implementation}{Implementation Details}
\saltappsubsectionentry{sec:main_results}{Main Results}
\saltappsubsectionentry{sec:ablation_results}{Ablation Studies and Analysis}
\saltappsectionentry{sec:conclusion}{Conclusion}
\vspace{0.5em}
{\footnotesize\itshape Appendix\par}
\vspace{0.15em}
\saltappsectionentry{app:implementation}{Detailed Implementation and Evaluation}
\saltappsubsectionentry{app:notation}{Notation}
\saltappsubsectionentry{app:evaluation_protocol}{Evaluation Protocol}
\saltappsubsectionentry{app:metrics}{Metric Definitions}
\saltappsubsectionentry{app:sampling}{Models and Sampling}
\saltappsubsectionentry{app:training}{Training Configuration}
\saltappsubsectionentry{app:high_resolution}{High-Resolution Configuration}
\saltappsectionentry{sec:visual_examples}{Additional Qualitative Results}
\saltappsubsectionentry{app:qualitative_480p}{480p Qualitative Comparisons}
\saltappsubsectionentry{app:guidance_provenance}{Additional Guidance Examples and Sample Provenance}
\saltappsubsectionentry{sec:hr_visuals}{High-Resolution Qualitative Comparisons}
\saltappsubsectionentry{app:visual_prompts}{Prompts for Main-Result Visualizations}
\end{minipage}%
}
\endgroup
\vspace{0.75em}

\section{Detailed Implementation and Evaluation}
\label{app:implementation}

\subsection{Notation}
\label{app:notation}

Tab.~\ref{tab:notation} collects the symbols used in Sec.~\ref{sec:method_csf}--\ref{sec:method_stage4} and in this appendix.

\begin{table}[h]
    \centering
    \small
    \setlength{\aboverulesep}{0.15ex}\setlength{\belowrulesep}{0.25ex}
    \renewcommand{\arraystretch}{1.05}
    \begin{tabularx}{\textwidth}{@{}l>{\raggedright\arraybackslash}X@{}}
        \toprule
        \multicolumn{2}{@{}l}{\textit{Blocks and contexts}} \\
        $K$ & Number of temporally aligned audio--video blocks \\
        $\vz_k=(\vz_k^v,\vz_k^a)$ & Video and audio latents of block $k$ \\
        $m\in\{v,a\}$ & Modality index \\
        $\vy$ & Text condition \\
        $\vc_k$ & Conditional information visible at block $k$ \\
        $\vc_k^\star$ & Clean teacher-forced context \\
        $\tilde{\vc}_k$ & Noise-mixed context seen by the CSF student \\
        $\hat{\vc}_k$ & Rollout context built from generated blocks \\
        $\vc_G$, $\vc_R$, $\vc_D$ & Contexts of generator sampling, real-score evaluation, fake-score training \\
        \midrule
        \multicolumn{2}{@{}l}{\textit{Flow path}} \\
        $t\in[0,1]$ & Noise level; larger $t$ is noisier \\
        $\vepsilon_k$ & Gaussian noise for block $k$ \\
        $\vz_{k,t}$ & Noisy target block $(1-t)\vz_k+t\vepsilon_k$ \\
        $\gamma_i$ & Noise level applied to history block $i<k$ in CSF \\
        $\tilde{\vz}_i^m$ & Noise-mixed history latent of modality $m$ \\
        \midrule
        \multicolumn{2}{@{}l}{\textit{Models}} \\
        $F_\eta$, $F_{\bar\eta}$ & CSF student and its EMA teacher \\
        $H_{\eta,m}^{\ell}$ & Layer-$\ell$ representation of modality $m$ \\
        $\ell_s<\ell_d$ & Shallow student layer and deep EMA-teacher layer \\
        $P_m$ & Two-layer projection head for modality $m$ \\
        $G_\theta$ & Few-step causal generator \\
        $R_\psi$, $D_\phi$ & Frozen real score and online fake score \\
        $U_\omega$ & Causal latent upsampler \\
        \midrule
        \multicolumn{2}{@{}l}{\textit{Distributions and predictions}} \\
        $p_{\theta,k}(\cdot\mid\vc_k)$ & Generator-induced conditional over clean blocks \\
        $p_{\theta,k,t}$, $p_{\mathrm r,k,t}$ & Perturbed generator and reference conditionals \\
        $\hat{\vz}_k^{G}$, $\hat{\vz}_k^{R}$, $\hat{\vz}_k^{D}$ & Clean predictions of generator, real score, fake score \\
        $\tilde{\vz}_{k,t}$ & Perturbed generated block used for scoring \\
        $\vg_k^m$ & Normalized DMD direction \\
        \midrule
        \multicolumn{2}{@{}l}{\textit{Objectives and weights}} \\
        $\mathcal{L}_{\mathrm{FM}}$, $\mathcal{L}_{\mathrm{rep}}^m$, $\mathcal{L}_{\mathrm{CSF}}$ & Flow-matching, representation-alignment, and total CSF losses \\
        $\mathcal{L}_{\mathrm{DMD},k}$, $\mathcal{L}_{\mathrm{fake}}$, $\mathcal{L}_G$ & Block-conditional DMD objective, fake-score loss, generator surrogate \\
        $\mathcal{L}_{\mathrm{scale}}$ & Scale-wise objective summed over the two scales \\
        $\lambda_a$, $\lambda_{\mathrm{rep}}$ & Audio weight and representation-alignment weight \\
        \midrule
        \multicolumn{2}{@{}l}{\textit{Scale-wise stage}} \\
        $\mathcal{S}_L$, $\mathcal{S}_H$ & Low- and high-resolution noise grids \\
        $\hat{\vz}_{1:K}^{L}$, $\hat{\vz}_{1:K}^{H}$ & Completed rollout at each scale \\
        \bottomrule
    \end{tabularx}
    \caption{\textbf{Notation.}}
    \label{tab:notation}
\end{table}

\subsection{Evaluation Protocol}
\label{app:evaluation_protocol}

We evaluate joint text-to-audio--video generation on the 1,000 prompts of
JavisBench-mini~\citep{liu2025javisdit}, covering diverse visual events and sound
sources. The \emph{official} track uses the released joint prompts and evaluator.
Every model generates one sample per prompt with seed $12{,}345+i$ for sample
index $i$; video and audio are evaluated from H.264 MP4 and PCM WAV outputs.
To test sensitivity to prompt formulation, we also construct a fixed
\emph{rewrite} track by applying the official LTX-2 prompt enhancer~\citep{hacohen2026ltx2}
once to each joint prompt with rewrite seed 42. The resulting 1,000 longer
prompts are shared across models, rather than rewritten separately for each
system.

\subsection{Metric Definitions}
\label{app:metrics}

The official main comparison reports visual quality (VQ), motion quality
(MQ), audio quality (AQ), CLIP text--video agreement, ImageBind audio--visual
agreement (IB-AV), JavisScore, and DeSync (lower is better). These distinguish
perceptual quality from semantic agreement and temporal synchronization.
We additionally report VBench aesthetic quality, imaging quality, subject
consistency, and background consistency on the same 480p generations.
These consistency measurements concern the evaluated clips, not extended
rollouts. The initialization analysis includes ImageBind text--video (IB-TV)
and text--audio (IB-TA) agreement, while the causal-teacher analysis also uses
AVHScore and CLAP. AV-Align is excluded because of its evaluation cost.
The rewrite track reports VQ, MQ, AQ, IB-AV, AVHScore, JavisScore, and DeSync;
we omit text-dependent metrics because the longer joint rewrites may exceed
the encoders' text contexts and do not provide separately rewritten audio
and video descriptions.

\paragraph{Interpreting DeSync under degraded video.}
DeSync estimates the temporal offset between an audio track and a video
track, without assessing the quality of either. A configuration whose video
is nearly static therefore offers little motion for the synchronization
model to misalign and can record a low DeSync while producing visually
uninformative output. This is the case for the BI--AR row of
Tab.~\ref{tab:stage2_context_ablation}(a): it attains the best DeSync in
that table at $0.531$, but reaches only $1.131$ VQ and $0.157$ MQ, far below
the aligned AR--AR configuration at $3.147$ and $1.351$. We therefore read
DeSync together with VQ and MQ rather than in isolation, and treat it as
informative only among configurations that produce comparable motion.

\subsection{Models and Sampling}
\label{app:sampling}

LTX-2 Base~\citep{hacohen2026ltx2} is the bidirectional foundation-model reference;
the CSF AR teacher measures causal generation before step reduction.
OmniForcing~\citep{su2026omniforcing} is the released 4-step causal baseline.
At 480p, Salt++ is evaluated through two complete routes, initialized by either
TF-dCM or context-aligned AR--AR TF-DMD and then refined on generated contexts.
Thus, a \emph{route} in the main comparison includes rollout refinement, whereas
an \emph{initializer} in the ablations does not.
We generate $832\times480$ videos with 121 frames at 24 FPS, approximately five
seconds. The LTX-2 VAE compresses video temporally by $8\times$, so the
block-causal layout pairs three video latent frames with 25 audio latent
frames per one-second block; the first block additionally carries the initial
latent frame of each stream, and a 121-frame clip therefore forms $K=5$
blocks. The AR teacher uses 40-step Euler sampling with video and audio CFG
both set to 4.0. The Salt++ initializers and rollout models use the 4-step
grid $[1000,960,889,727,0]$ and inference CFG $=1$, predicting a clean endpoint
and re-noising it to the next grid point at each step. Both this grid and the
finer eight-point grid from which training noise levels are drawn
(Appendix~\ref{app:training}) are images of uniform partitions of $[0,1]$
under the same shift $t\mapsto8t/(1+7t)$ used during training,
so the former is a subset of the latter. At 960p, the scale-wise
system uses 2 low-resolution and 2 high-resolution generator evaluations.
LTX-2 uses its $40+3$-step high-resolution path. OmniForcing is evaluated by
directly applying its released checkpoint to the 960p token grid; this is a
resolution-extrapolation diagnostic, not a resolution-matched trained baseline.

\subsection{Training Configuration}
\label{app:training}

All post-training stages use internal audio--video data at $832\times480$ and
24 FPS, the block layout of Appendix~\ref{app:sampling}, mixed precision, and
FSDP, with one sample per device.

\paragraph{Causal Self-Flow.}
Training starts from LTX-2 and keeps the velocity parameterization of
Eq.~\ref{eq:conditional_fm}. Target noise levels are drawn uniformly on
$[0.003,1]$ and reshaped by the shift $t\mapsto8t/(1+7t)$, which
concentrates supervision at higher noise; per-sample losses
are weighted by $w(t)=\exp(-2(t-0.5)^2)$, which de-emphasizes both
extremes. History noise levels are drawn from $\mathcal{U}(0,0.5)$, and the
student layer $\ell_s=14$ is aligned to EMA-teacher layer $\ell_d=34$ of the
48-layer backbone. Projection heads are \texttt{Linear--SiLU--Linear} with a
zero-initialized final layer, and the EMA decay is $0.99$. We set
$\lambda_a=0.16$ and $\lambda_{\mathrm{rep}}=0.05$, and alternate CSF and clean
teacher-forcing updates with equal probability. Optimization uses AdamW at
learning rate $1\times10^{-4}$, dropped to $5\times10^{-5}$ after 6k
iterations, with 100 warmup steps, weight decay $0.01$, and gradient clipping
at $10.0$, on 32 GB300 GPUs.

\paragraph{Context-aligned AR DMD.}
The generator, the frozen real score, and the online fake score are all
initialized from the same CSF checkpoint, so the real score is exactly the AR
teacher of Sec.~\ref{sec:method_csf}. Each iteration evaluates the generator
once, at a single entry noise level drawn uniformly from the eight non-zero
points of $[1000,982,960,930,889,828,727,533,0]$ and shared across all blocks
and both modalities; the four levels visited at inference are among them.
Score noise levels for the DMD update and for
fake-score training are drawn independently from the same shifted-uniform
distribution, supported on $[0.02,0.98]$, and all blocks contribute
equally to the loss. The fake score is updated every iteration and the
generator every five. Unless ablated, video and audio teacher guidance are
sampled independently from $\mathcal{U}(1.0,3.5)$ against a fixed negative
prompt; these are training-time guidance values, distinct from inference CFG.
We train for 3.2k iterations on 12 GB300 GPUs, using AdamW at learning rate
$2\times10^{-5}$ for the generator and $5\times10^{-5}$ for the fake score,
with 100 warmup steps, weight decay $0.01$, and gradient clipping at $1.0$.

\paragraph{TF-dCM comparison route.}
The TF-dCM student and its frozen causal teacher are initialized from the
same CSF checkpoint as the DMD route, so the two routes differ only in the
distillation objective. Following Causal-rCM~\citep{zheng2026causalrcm}, we
use dense adjacent-pair discrete consistency distillation over a 48-point
discretization with unit skipping interval and a consistency loss scale of
$100$, under the same shift and $[0.003,1]$ noise range as CSF. Every
teacher bridge step applies fixed video and audio guidance of $4.0$ against
the same negative prompt; this is the setting that
Sec.~\ref{sec:stage2_results} compares against randomized lower guidance.
Only the student is optimized, with no fake score and no EMA target. We train
for 4.8k iterations on 12 GB300 GPUs, using AdamW with
$(\beta_1,\beta_2)=(0,0.999)$ as in the Causal-rCM recipe, learning rate
$2\times10^{-5}$, 100 warmup steps, weight decay $0.01$, and gradient
clipping at $10.0$.
The loss scale is worth singling out. The dCM objective produces very small
values, typically around $10^{-4}$, and we found training with the unscaled
loss to be substantially worse than with the factor of $100$. Adam is
invariant to a constant factor on the loss in exact arithmetic, so we
attribute the difference to behaviour at this magnitude: the default
$\epsilon=10^{-8}$ is no longer negligible against the second-moment
estimate, and mixed-precision gradients lose relative precision. We therefore
retain the Causal-rCM scale rather than treating it as a free
hyperparameter.

\paragraph{On-policy continuation.}
Both routes then continue on generated histories with a shared self-forcing
setup: the generator rolls out under the 4-step inference grid with a KV
cache, one exit is sampled per iteration and shared across all blocks, and
the fake score is trained on rollouts drawn from the same exit distribution.
This stage consumes prompts only; no paired video or audio is loaded. Score
noise levels, the five-to-one fake/generator update ratio, and the uniform
block weighting match the clean-prefix stage. The two routes differ in what
they can carry forward. The AR--AR route inherits both its generator and its
fake score from the clean-prefix AR--AR checkpoint and keeps the causal CSF
teacher as the real score, so the context-aligned objective continues
unchanged for 1.2k iterations with teacher guidance still drawn from
$\mathcal{U}(1.0,3.5)$. The TF-dCM route has no fake score to inherit, so
both score models are initialized from LTX-2 and evaluated bidirectionally,
giving the BI--BI configuration used by prior recipes; it runs for 2.4k
iterations with guidance fixed at $3.0$. Optimization in both cases uses
AdamW at learning rate $2\times10^{-5}$ for the generator and
$5\times10^{-5}$ for the fake score, with a cosine schedule, 100 warmup
steps, weight decay $0.01$, and gradient clipping at $10.0$, on 12 GB300
GPUs.

\paragraph{Ablation settings.}
The score-context ablations use controlled training configurations and the
same evaluation protocol across variants. For the teacher-guidance comparison
in Tab.~\ref{tab:stage2_context_ablation}(b), both AR--AR TF-DMD models are
evaluated at 3.2k training iterations, using 4-step inference with CFG $=1$
and the same prompts and sample-index seeds. Video and audio teacher guidance
are either both fixed at 4.0 or sampled independently from
$\mathcal{U}(1.0,3.5)$.

\subsection{High-Resolution Configuration}
\label{app:high_resolution}

Scale-wise post-training uses $\mathcal{S}_L=[1.0,0.960,0]$ and
$\mathcal{S}_H=[0.889,0.727,0]$, connected by causal $2\times$ latent upsampling.
Training samples
1- or 2-call exits independently at each scale, shared across temporal
blocks, and weights the two scale losses equally. As described in
Sec.~\ref{sec:method_stage4}, the generator is causal while the scale-wise
score models evaluate completed rollouts bidirectionally; the upsampler
receives HR gradients when the sampled node directly consumes its output.
All four generator calls are executed at inference to produce
$1664\times960$ videos. Call counts describe the sampling budget, not a
measurement of equal wall-clock cost across resolutions or systems.

\paragraph{Initialization and schedule.}
The scale-wise generator starts from the AR--AR DMD route. Because this stage
scores rollouts bidirectionally, we first warm up that generator for 1.6k
iterations of 480p on-policy DMD under bidirectional real and fake scores,
matching the scoring configuration used by scale-wise training, and then run
2.4k scale-wise iterations. The fake score is carried over from the warm-up,
the real score stays frozen, and $U_\omega$ starts from the 1.2k checkpoint
described below and is updated throughout. Teacher guidance is fixed at $3.0$
for video and $5.0$ for audio. Optimization uses AdamW at learning rate
$2\times10^{-5}$ for the generator and $5\times10^{-5}$ for the fake score,
with a cosine schedule, 100 warmup steps, weight decay $0.01$, and gradient
clipping at $10.0$, on 12 GB300 GPUs.

\paragraph{Causal upsampler initialization.}
The causal upsampler $U_\omega$ retains the architecture and weights of the
released LTX-2 spatial upscaler and becomes causal through input windowing:
to produce block $[s,e)$ it runs the network on latent frames
$[\max(0,s-18),e)$ and crops the current block from the result. Restricting
the receptive field in this way degrades the output, and we recover it by
imitation. The teacher is the same released upscaler applied to the entire
low-resolution sequence at once; the loss is a per-frame squared error in
normalized latent space between each causal block output and the
corresponding frames of the teacher output, averaged over non-padded frames.
The inputs are online low-resolution rollouts from the frozen 4-step 480p
generator rather than ground-truth latents, so $U_\omega$ is distilled on the
latent distribution it encounters at deployment; each rollout draws its
sampling grid with equal probability from the 4-call grid used at inference
and from a finer 8-call grid. We train for 1.2k iterations with AdamW at
learning rate $2\times10^{-5}$, a cosine schedule with 100 warmup steps,
gradient clipping at $1.0$, and mixed precision, on the same training data
as the preceding stages. This procedure only
initializes $U_\omega$; scale-wise post-training then continues to update it
through the high-resolution term of Eq.~\ref{eq:stage4_scale_loss}.

\section{Additional Qualitative Results}
\label{sec:visual_examples}

\subsection{480p Qualitative Comparisons}
\label{app:qualitative_480p}

\begin{figure}[t]
    \centering
    \includegraphics[width=\linewidth]{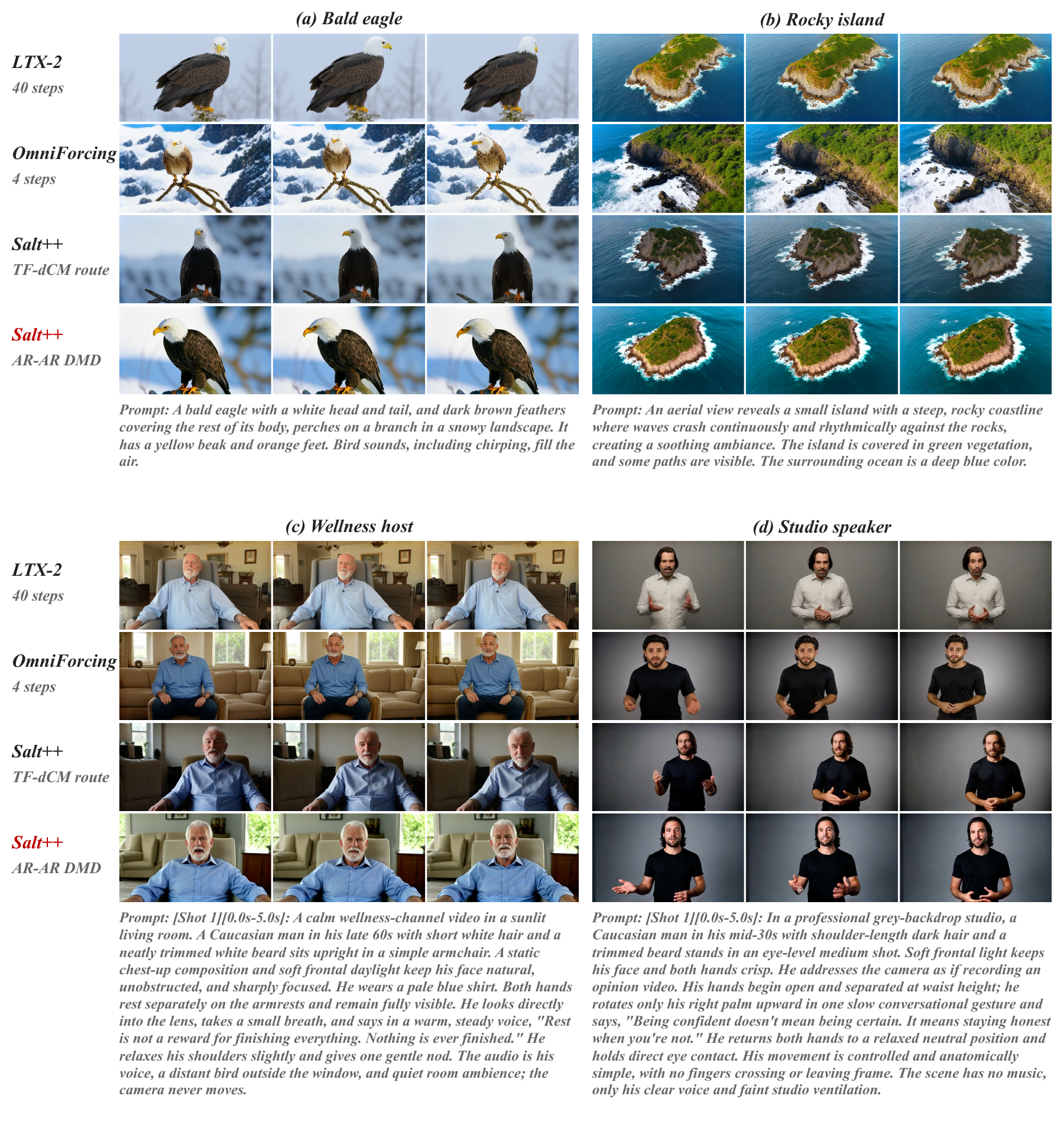}
    \caption{\textbf{Additional 480p comparisons.}
    (a) Bald eagle, (b) Rocky island, (c) Wellness host, and (d) Studio speaker.
    Each case uses the method order of Fig.~\ref{fig:main_coast_diner}.}
    \label{fig:appendix_480p_set1}
\end{figure}

\begin{figure}[t]
    \centering
    \includegraphics[width=\linewidth]{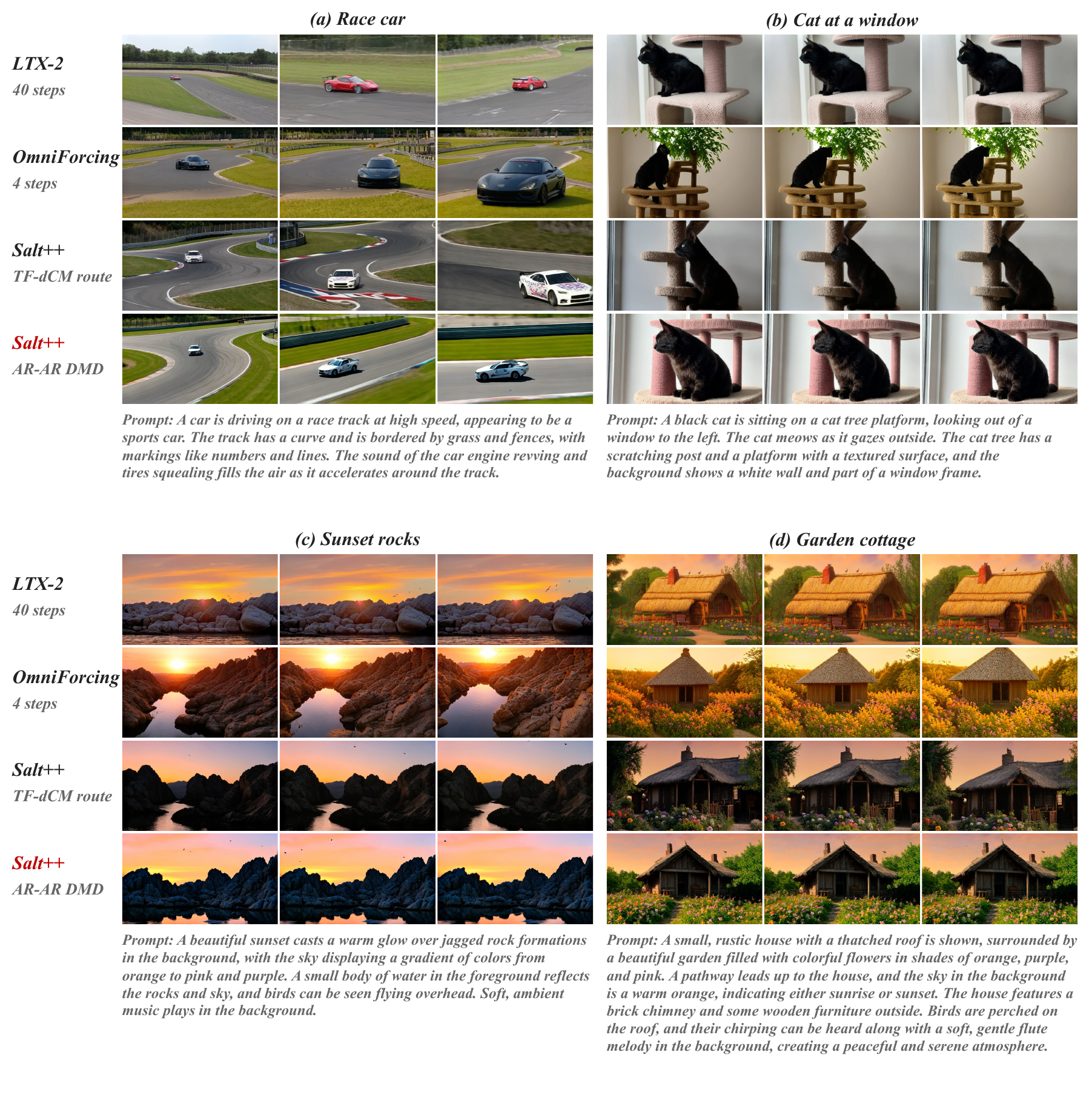}
    \caption{\textbf{Additional 480p comparisons.}
    (a) Race car, (b) Cat at a window, (c) Sunset rocks, and (d) Garden cottage.
    Method ordering follows Fig.~\ref{fig:main_coast_diner}.}
    \label{fig:appendix_480p_set2}
\end{figure}

\paragraph{Presentation protocol.}
The 480p appendix comparisons contain four cases per figure, arranged in two
pairs, with three frames per case. All methods share the prompt and displayed
times within each case; frames are not retouched or color corrected.
LTX-2 Base uses 40 steps, while OmniForcing and both Salt++ routes use four.
Figs.~\ref{fig:appendix_480p_set1} and~\ref{fig:appendix_480p_set2}
complement Fig.~\ref{fig:main_coast_diner} with eight additional examples.

\clearpage

\clearpage

\subsection{Additional Guidance Examples and Sample Provenance}
\label{app:guidance_provenance}

\begin{figure}[t]
    \centering
    \includegraphics[width=0.98\linewidth]{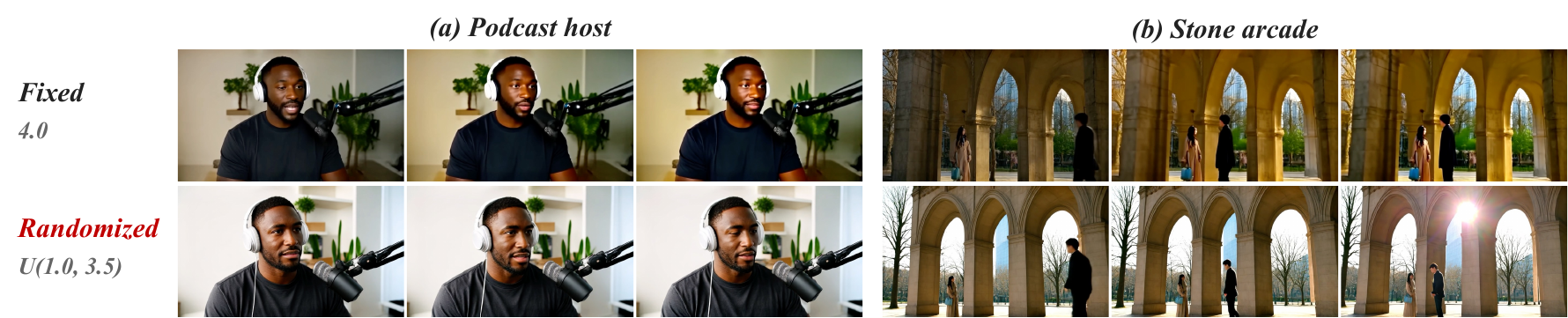}
    \caption{\textbf{Training-time guidance calibration: podcast and arcade examples.}
    The lower randomized setting preserves more tonal and texture detail
    than fixed high guidance.}
    \label{fig:guidance_qualitative}
\end{figure}

\begin{figure}[t]
    \centering
    \includegraphics[width=\linewidth]{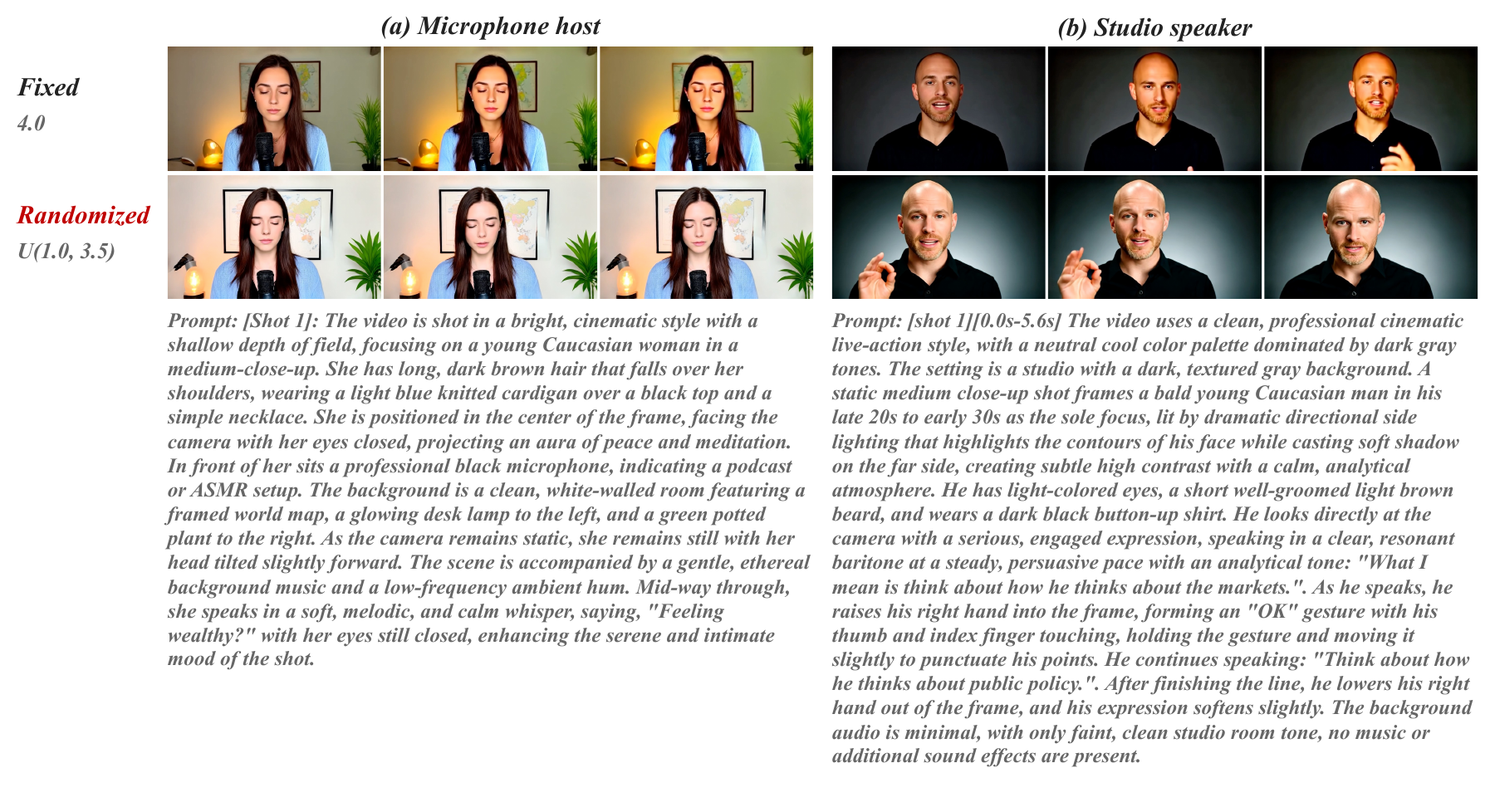}
    \caption{\textbf{Additional training-guidance comparisons: Microphone host and Studio speaker.} The two AR--AR TF-DMD checkpoints use the same settings as Fig.~\ref{fig:guidance_qualitative}: fixed $4.0$ versus independent $\mathcal{U}(1.0,3.5)$ teacher guidance at 3.2k, 4-step inference, and CFG $=1$.}
    \label{fig:guidance_appendix}
\end{figure}

\paragraph{Setup for the qualitative comparisons.}
The TF-dCM route in the main comparison and the appendix cases uses the
Stage~3 2.4k checkpoint, whereas TF-dCM in the initialization ablation is the
Stage~2 4.8k model. Both ablation figures use 4-step inference with CFG $=1$
and matched prompts, sample-index seeds, and frame times. The guidance
examples come from a separate internal visualization set rather than the
1,000-prompt quantitative evaluation; Fig.~\ref{fig:guidance_appendix} adds
two further speaking cases under the same settings. Matched prompts and seeds
do not imply pixel-aligned compositions.

\clearpage
\subsection{High-Resolution Qualitative Comparisons}
\label{sec:hr_visuals}

These examples use matched prompts, sample-index seeds,
and times: Fitness coach (0007, 4.50\,s) and Wellness host (0017, 0.50\,s).
Both follow the four-column layout of Fig.~\ref{fig:hr_fitness}, with full
frames above and marked detail windows below.

\begin{figure}[t]
    \centering
    \includegraphics[width=0.95\linewidth]{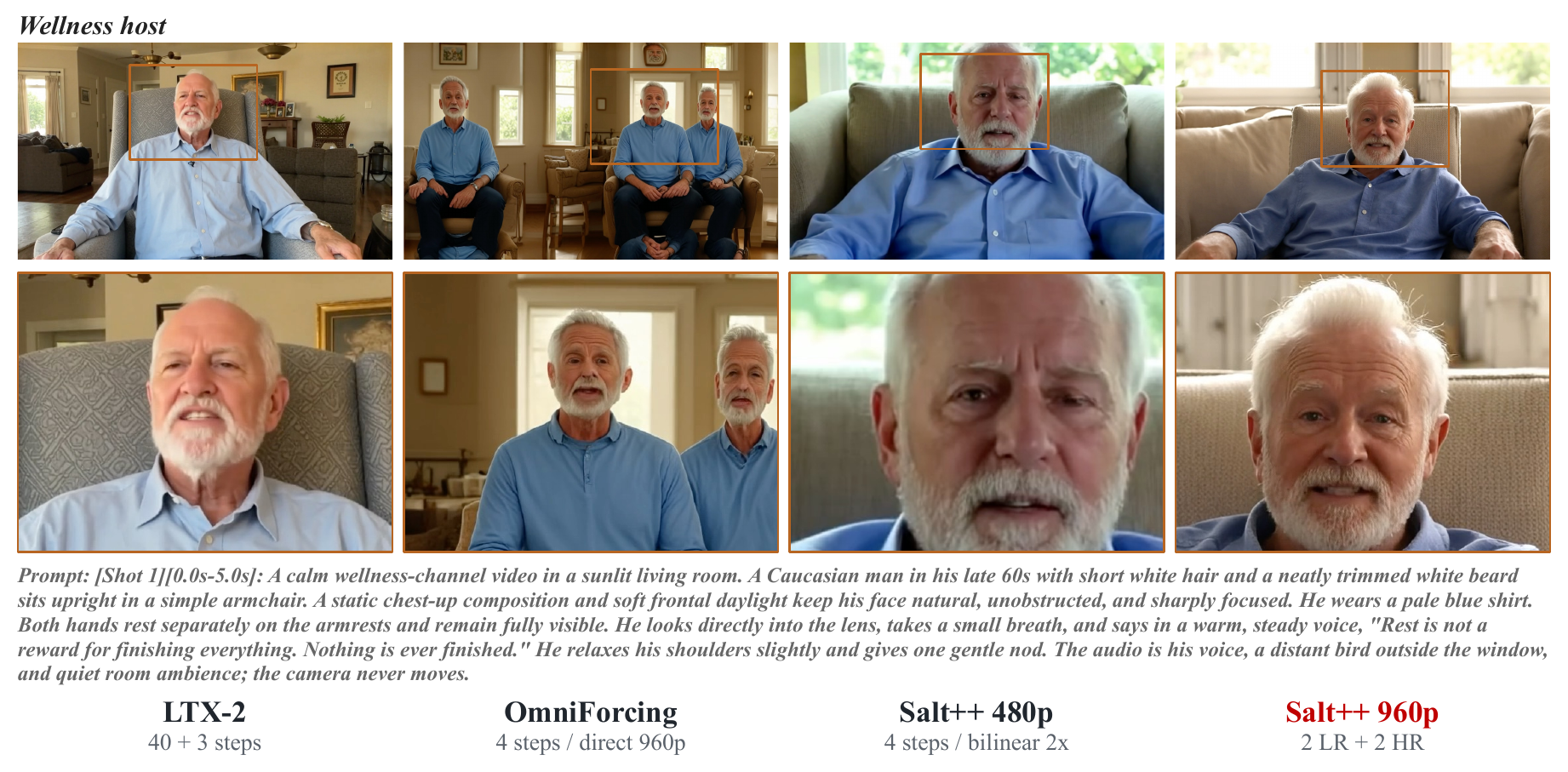}
    \caption{\textbf{Additional 960p comparison: Wellness host.}
    Full frames above and the corresponding facial and hair detail windows
    below, using the four-method column order of Fig.~\ref{fig:hr_fitness}.}
    \label{fig:hr_appendix}
\end{figure}

\paragraph{Checkpoints and display protocol.}
These qualitative examples use the Stage~3 SALT BI--BI 4.0k parent and its
Stage~4a 3.2k scale-wise output, without extra refiner or DPO; this parent is
not the AR--AR rollout checkpoint used in the 480p main figure. They are
selected visualizations, separate from the 1,000-prompt aggregate evaluation.
The 480p parent is bilinearly upsampled for display; the native 960p model uses
2 low-resolution and 2 high-resolution generator calls. All columns show
matched times and equal-sized detail windows.
OmniForcing uses the released causal model on the 960p token grid with four
updates. Its repeated spatial patterns are a resolution-extrapolation result,
not a statement about performance at its training resolution. Detail windows
cover 34\% of frame width and 44\% of frame height in every column; their
positions follow the subject without face-size normalization. These are
independent generations, not pixel-aligned super-resolution pairs. No frame
is retouched, sharpened, or color corrected. Selected stills do not establish
full-video temporal or audio quality.

\subsection{Prompts for Main-Result Visualizations}
\label{app:visual_prompts}

We provide the full input prompts for the main-result examples in
Figs.~\ref{fig:main_coast_diner} and~\ref{fig:hr_fitness} below.
Prompts for the appendix examples are shown beneath each case in
Figs.~\ref{fig:appendix_480p_set1}--\ref{fig:hr_appendix} and are not repeated here.
These are the original generation inputs, shared by all methods within each
case. Sampling settings are recorded in
Appendices~\ref{app:guidance_provenance} and~\ref{sec:hr_visuals}.

\paragraph{White-cliff beach.}
JavisBench 0631, Fig.~\ref{fig:main_coast_diner}.

\begin{quote}
\small
A beautiful beach with large white cliffs on either side and a sandy shoreline is shown. The cliffs have a rough texture with some greenery visible, and ocean waves crash against the rocks at their base. Sunlight filters through partly cloudy skies, casting a warm and serene glow. The sound of waves crashing against the rocks and the gentle breeze blowing contributes to the peaceful atmosphere.
\end{quote}

\paragraph{Diner conversation.}
Fig.~\ref{fig:main_coast_diner}.

\begin{quote}
\small
[Shot 1][0.0s-5.0s]: A moody cinematic diner at night, seen in a locked medium close-up across a booth. A Caucasian woman in her early 40s with a short dark bob wears a charcoal coat and sits beneath soft amber light. Rain streaks the window behind her and a red neon glow remains blurred outside. Her face stays unobstructed and sharply focused. A white coffee cup sits near her left hand; both hands rest separately on the table with all fingers visible. She looks toward an unseen person opposite her, slowly tightens her fingertips around the cup without lifting it, and says in a low controlled voice, "You weren't supposed to find that letter." Her eyes hold steady after the line. Rain, a distant refrigerator hum, and her voice are the only sounds.
\end{quote}

\paragraph{Fitness coach.}
Fig.~\ref{fig:hr_fitness}.

\begin{quote}
\small
[Shot 1][0.0s-5.0s]: In a minimalist fitness studio with a neutral grey wall, a mature Black man with a shaved head and a short grey beard stands on a black mat. The camera is locked in an eye-level medium shot, and broad soft lighting keeps his face, shoulders, and hands crisp without motion blur. He wears a maroon athletic shirt. Both hands begin open at waist height, palms angled inward and clearly separated. He brings them slowly upward by a few inches while maintaining direct eye contact and says in a calm, supportive voice, "You don't need more motivation; you need one smaller promise." His hands stop and remain still as he gives a gentle nod. The studio is quiet except for his resonant voice and faint ventilation hum.
\end{quote}

\end{document}